\documentclass[sigconf]{acmart}

\copyrightyear{2025}
\acmYear{2025}
\setcopyright{acmlicensed}\acmConference[CIKM '25]{Proceedings of the 34th ACM International Conference on Information and Knowledge Management}{November 10--14, 2025}{Seoul, Republic of Korea}
\acmBooktitle{Proceedings of the 34th ACM International Conference on Information and Knowledge Management (CIKM '25), November 10--14, 2025, Seoul, Republic of Korea}
\acmDOI{10.1145/3746252.3761381}
\acmISBN{979-8-4007-2040-6/2025/11}

\usepackage{amsmath,amsfonts,bm}

\def\eqref#1{equation~\ref{#1}}
\def\1{\bm{1}}

\def\vc{{\bm{c}}}

\def\vh{{\bm{h}}}

\def\vp{{\bm{p}}}

\def\vr{{\bm{r}}}

\def\vv{{\bm{v}}}
\def\vw{{\bm{w}}}
\def\vx{{\bm{x}}}
\def\vy{{\bm{y}}}

\def\mH{{\bm{H}}}

\def\mW{{\bm{W}}}
\def\mX{{\bm{X}}}

\DeclareMathAlphabet{\mathsfit}{\encodingdefault}{\sfdefault}{m}{sl}
\SetMathAlphabet{\mathsfit}{bold}{\encodingdefault}{\sfdefault}{bx}{n}

\usepackage{hyperref}
\usepackage{url}
\usepackage{graphicx}
\usepackage{natbib}
\usepackage{booktabs}
\usepackage{amsmath}
\usepackage{subcaption}
\usepackage{amsfonts} 
\usepackage{enumitem}
\usepackage{colortbl}
\usepackage{dsfont}
\usepackage{xcolor}
\usepackage[most]{tcolorbox}
\usepackage{booktabs}
\usepackage{wrapfig,lipsum,booktabs}

\usepackage{array}

\usepackage{multirow}
\usepackage{pifont}

\usepackage{enumitem}

\usepackage{setspace}
\usepackage{algorithm}
\usepackage{algpseudocode}
\usepackage{subcaption}
\usepackage{caption}
\usepackage[most]{tcolorbox} 
\begin{document}
\title{ProtoEHR: Hierarchical Prototype Learning for EHR-based Healthcare Predictions}
\author{Zi Cai}
\orcid{0009-0001-0846-629X}
\authornote{Work done while Zi Cai was a research intern at The University of Oxford.}
\affiliation{%
  \institution{University of Cambridge}
  \city{Cambridge}
  \country{United Kingdom}
}
\email{caizicharlesofficial@gmail.com}

\author{Yu Liu}
\orcid{0000-0002-2399-2829}
\authornote{Yu Liu is the corresponding author.}
\affiliation{%
  \institution{University of Oxford}
  \city{Oxford}
  \country{United Kingdom}}
\email{yu.liu@eng.ox.ac.uk}

\author{Zhiyao Luo}
\orcid{0000-0003-0015-2023}
\affiliation{%
  \institution{University of Oxford}
  \city{Oxford}
  \country{United Kingdom}}
\email{zhiyao.luo@eng.ox.ac.uk}

\author{Tingting Zhu}
\orcid{0000-0002-1552-5630}
\affiliation{%
  \institution{University of Oxford}
  \city{Oxford}
  \country{United Kingdom}}
\email{tingting.zhu@eng.ox.ac.uk}

\begin{abstract}
Digital healthcare systems have enabled the collection of mass healthcare data in electronic healthcare records (EHRs), allowing artificial intelligence solutions for various healthcare prediction tasks. However, existing studies often focus on isolated components of EHR data, limiting their predictive performance and interpretability. To address this gap, we propose ProtoEHR, an interpretable hierarchical prototype learning framework that fully exploits the rich, multi-level structure of EHR data to enhance healthcare predictions. More specifically, ProtoEHR models relationships within and across three hierarchical levels of EHRs: medical codes, hospital visits, and patients. We first leverage large language models to extract semantic relationships among medical codes and construct a medical knowledge graph as the knowledge source. Building on this, we design a hierarchical representation learning framework that captures contextualized representations across three levels, while incorporating prototype information within each level to capture intrinsic similarities and improve generalization. To perform a comprehensive assessment, we evaluate ProtoEHR in two public datasets on five clinically significant tasks, including prediction of mortality, prediction of readmission, prediction of length of stay, drug recommendation, and prediction of phenotype. The results demonstrate the ability of ProtoEHR to make accurate, robust, and interpretable predictions compared to baselines in the literature. Furthermore, ProtoEHR offers interpretable insights on code, visit, and patient levels to aid in healthcare prediction.
\end{abstract}




\maketitle

\vspace{-0.3cm}
\section{Introduction}
The digitization of healthcare systems in recent years has led to the accumulation of substantial electronic health records (EHRs) \citep{shickel2017deep,sarwar2022secondary}. The patient's EHR data contain demographics, as well as detailed information about the hospital visit, including medical procedures performed, clinical diagnoses, and prescribed medications. Numerous studies \citep{wang2020mimic,jiang2023graphcare} have focused on developing models to facilitate the prediction of patient outcomes, such as mortality risk \citep{choi2018mime} and readmission possibilities \citep{wang2021predictive}, as well as personalized treatment strategies, including classification of phenotypes \citep{harutyunyan2019multitask} and drug recommendation \citep{shang2019gamenet}. Accomplishing these tasks assists clinical decision-making, improves treatment quality, and allows better allocation of resources.

Existing frameworks for healthcare prediction capitalize on the various characteristics of EHR data \cite{sarwar2022secondary}. Each patient can have multiple hospital visits at different times. Each hospital visit is for specific reasons, and with doctors reaching varying diagnoses, ordering the required procedures, and prescribing the corresponding medications (typically recorded using a standardized coding system, for example, ICD codes\footnote{\url{https://www.who.int/standards/classifications/classification-of-diseases}}). Therefore, EHR data naturally form a \textit{hierarchical structure} that spans the levels of the patient, the visit, and the code \cite{choi2016retain,choi2018mime}. This structure provides a rich and interpretable representation of a patient’s clinical journey, enabling models to capture both temporal and semantic dependencies across levels for more effective healthcare prediction.

Beyond hierarchical connections across different levels of EHR data, it is equally important to consider the \textit{intrinsic similarities} among entities within each level. At the patient level, individuals may share lifestyle habits that influence their risk factors, increasing the propensity towards certain illnesses, e.g., smoking and poor diet increase the likelihood of coronary heart disease \cite{kannel1961factors}. At the visit level, patients may visit hospitals for similar purposes. It could be a routine visit to pick up prescriptions or it could be an emergency. Thus, we argue that visits also exhibit similarities with predictive value. At the code level, medical codes are typically categorized into diagnoses, procedures, and medications \cite{johnson2016mimic}, suggesting that similarity can be inferred within each category. In addition, codes can also be examined from a pathological or toxicological perspective, further enriching their contextual interpretation. While prior works have leveraged patient-level similarities through prototype learning \cite{zhang2021grasp,zhu2024prism}, intrinsic similarities within the visit and code levels remain largely underexplored. We argue that modeling both within-level similarities and cross-level hierarchies offers significant potential to improve healthcare predictions by fully capturing the complex semantics in EHR data.

To effectively capture both the hierarchical structure and the intrinsic similarities within EHR data, we propose ProtoEHR, a hierarchical prototype learning framework for healthcare predictions. We first apply large language models (LLMs) to construct a medical knowledge graph (KG) that captures the rich semantic relationships among medical codes, thereby enhancing the code-level modeling. Subsequently, building on the medical KG, we develop three specialized local encoders to model information at the code, visit, and patient levels, in alignment with the natural structure of EHR data. At each level, we introduce a novel prototype-based encoder to capture intrinsic similarities among objects and enhance the quality of representation through shared information. The patient-level representation, together with the learned prototypes from all three levels, is then fed into a customized hierarchical fusion module that integrates information across levels. The output of this module is finally passed through a linear projection layer to generate task-specific predictions. This architecture not only enhances predictive performance across clinical tasks, but also improves interpretability by highlighting the relative contributions of code-, visit-, and patient-level information to corresponding tasks. In summary, our contributions are as follows:

\begin{itemize}[leftmargin=10px]
  \item We propose ProtoEHR, a novel hierarchical prototype learning framework that integrates prototype learning and hierarchical representation learning to model both within-level similarities and cross-level hierarchies in EHR data.
  \item We develop an interpretable mechanism based on hierarchical prototypes and conduct both qualitative and quantitative analyses to provide valuable insights into how different levels of the hierarchy contribute to clinically meaningful outcomes.
  \item We conduct extensive experiments on two real-world datasets across five prediction tasks, where ProtoEHR consistently achieves strong performance, outperforming state-of-the-art baselines.
\end{itemize}

\section{Preliminaries}

Here we introduce the mathematical definition of the EHR dataset and medical KG, as well as the research problem.

\textbf{Definition 1 (EHR Dataset).} An EHR dataset $\mathcal{D}=(\mathcal{C},\mathcal{V})$ is composed of a set of medical codes $\mathcal{C}=\{c_i\}_{i=1}^{N_{\text{code}}}$ and hospital visit information $\mathcal{V}=\{\mathcal{V}_i\}_{i=1}^{N}$, where $N_{\text{code}}$ and $N$ are the total numbers of unique medical codes and patients in the dataset, respectively. For each patient $i$, $\mathcal{V}_i$ is an ordered sequence of hospital visits, defined as $\mathcal{V}_i=(\mathcal{V}_{i,1}, \cdots, \mathcal{V}_{i, |\mathcal{V}_i|})$, where $|\mathcal{V}_i|$ denotes the number of visits. Each visit $\mathcal{V}_{i,j}$ contains multiple medical codes recorded during that visit, denoted as $\mathcal{V}_{i,j}=\{c_{i,j,k}\in\mathcal{C}\}_{k=1}^{|\mathcal{V}_{i,j}|}$, where $|\mathcal{V}_{i,j}|$ denotes the number of medical codes associated with the visit, and $c_{i,j,k}$ denotes the $k$-th medical code recorded in the $j$-th visit of patient $i$.

\textbf{Definition 2 (Medical KG).} A medical KG is defined as $\mathcal{G}=(\mathcal{C}, \mathcal{R},\mathcal{F})$, where $\mathcal{C}$ and $\mathcal{R}$ are the sets of entities and relations, respectively. The fact set is given by $\mathcal{F}=\{(c_{h}, r, c_{t})|c_{h},c_{t}\in\mathcal{C}, r\in\mathcal{R}\}$, where each triplet $(c_{h}, r, c_{t})$ indicates that a medical code $c_h$ is related to another one $c_t$ through relation $r$.

\textbf{Problem 1 (EHR-based Healthcare Prediction).} Given an EHR dataset $\mathcal{D}=(\mathcal{C},\mathcal{V})$, the objective of EHR-based healthcare prediction is to learn a patient-specific representation and predict the corresponding clinical outcome. For each patient $i$ with historical records $\mathcal{V}_i$, the prediction is performed using a function/model $f$, formulated as $\hat{y}_i=f(\mathcal{D}, i, \mathcal{V}_i)$, where $\hat{y}_i$ represents the predicted outcome. The nature of $\hat{y}_i$ varies depending on the specific prediction task, e.g., $\hat{y}\in\{0,1\}$ for mortality prediction, and $\hat{y}\in\{1,\cdots,K\}$ for length-of-stay prediction with $K$ discrete classes.

\begin{figure*}[htbp]
    \includegraphics[width=.9\linewidth]{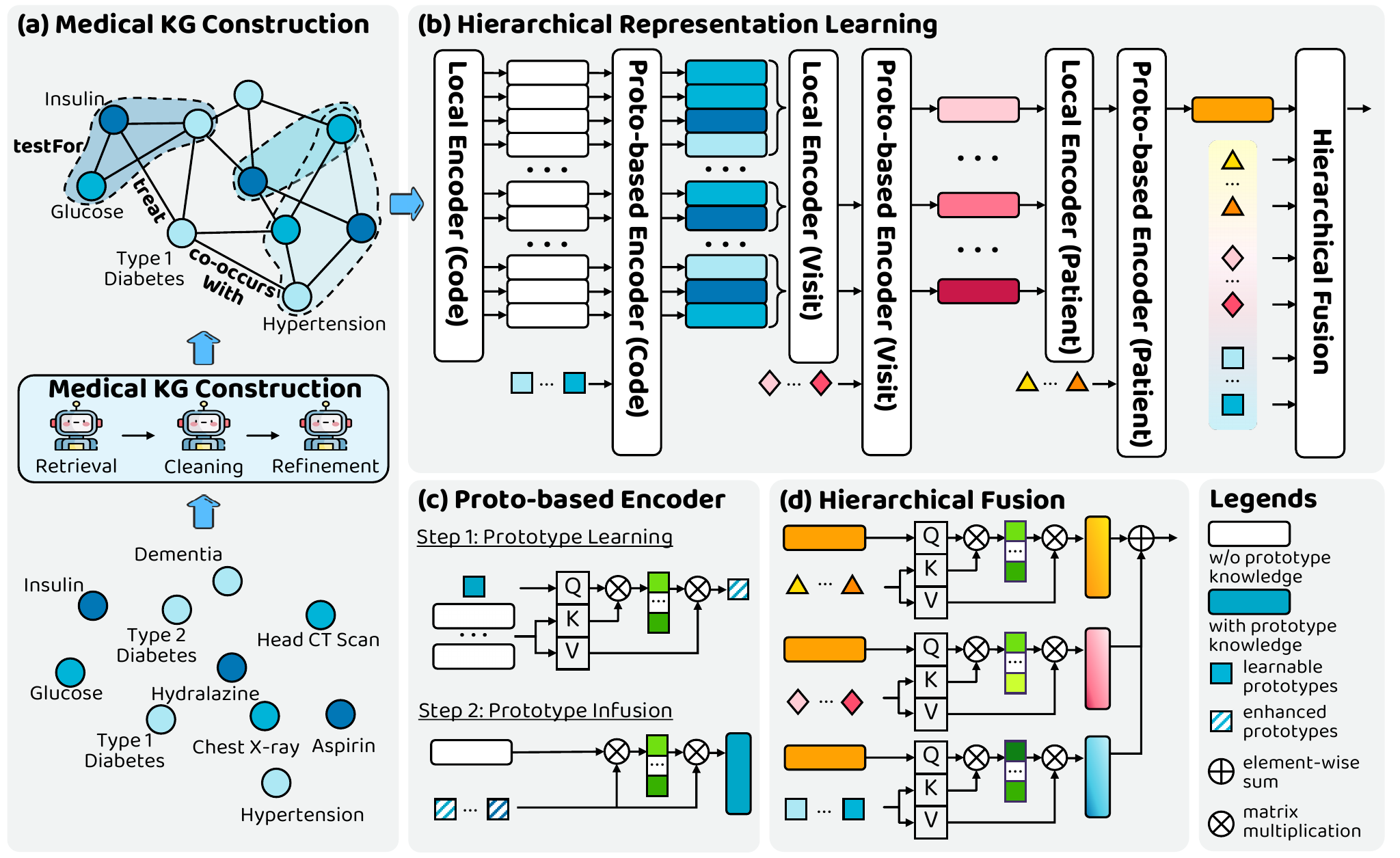}
    \vspace{-10px}
    \caption{The architecture of our proposed ProtoEHR framework. (a) A set of medical codes in EHR data is used to construct a medical KG with LLMs via three stages: retrieval, cleaning, and refinement. (b) Taking the constructed medical KG as well as patient information as input, hierarchical representations at the code, visit, and patient levels are learned. The local encoders take the representation of the previous level and learns a prototype-free representation. This is then used as input to the proto-based encoder to infuse prototype knowledge into the representation. (c) Prototype-based encoder consists of prototype learning and prototype infusion, where enhanced prototypes are first obtained and the representation is updated with these enhanced prototypes. (d) Hierarchical fusion aggregates the learned patient representation with hierarchical prototype knowledge. ``Proto-based'' is the abbreviation for ``prototype-based''.}
    \label{fig:architecture}
\vspace{-0.3cm}
\end{figure*}

\section{Proposed Method}
Figure~\ref{fig:architecture} presents an overview of our proposed ProtoEHR framework. It consists of three main stages: building the medical KG to exploit the relationships between medical codes in the EHR data; learning hierarchical representations and incorporating shared information with prototypes; and fusing the extracted patient representations with the learnt hierarchical prototypes for the prediction of healthcare outcomes. The structure and details of each of these stages are discussed below.

\subsection{Medical Knowledge Graph Construction}
To take advantage of the rich medical knowledge stored within the relationships between the medical codes, we constructed a medical KG with the diagnosis, procedure, and prescription codes in the dataset, as shown in Figure~\ref{fig:architecture}a. Since the numerous meaningful relationships therein are overwhelming for human experts to identify, we leverage powerful LLMs for automatic construction \cite{jiang2023graphcare,zhu2024llms}. 

Existing biomedical KGs, such as the UMLS-KG \cite{bodenreider2004unified}, could be used to incorporate medical knowledge but there is a possible mismatch between its ontologies and the medical code set $\mathcal{C}$, thus requiring further processing. Closed-source LLMs like GPT-4 \cite{achiam2023gpt} demonstrate strong capabilities in automatically retrieving triplets for KG construction but are prohibitively expensive when dealing with large entity sets, such as the medical code set in our case. In contrast, open-source LLMs, such as the Llama series \cite{dubey2024llama}, offer a more cost-effective alternative but have weaker retrieval capabilities. To balance quality and efficiency, we employ an open-source LLM to retrieve triplets, a closed-source LLM to train a classifier for filtering out false triplets, and a clustering-based approach to refine the relations. This hybrid strategy ensures that the constructed medical KG remains both expressive and robust. The three steps for KG construction are detailed as follows:

\paragraph{\emph{\textbf{Retrieval.}}} To begin with, we use an open-source LLM (Llama3-70B \cite{dubey2024llama}) denoted as $LLM_1$ to discover the semantic relationships within the medical codes. Taking each pair of the medical codes in the medical code set $\mathcal{C}$ as input, $LLM_1$ identifies the set of plausible triplets, $\mathcal{F}_1$. Denoting the pair of codes as $\texttt{ENTITY}_1$ and $\texttt{ENTITY}_2$, the extraction process can be expressed as:

$\text{Prompt}_1(\texttt{ENTITY}_1, \texttt{ENTITY}_2)\xrightarrow{LLM_1} (\texttt{ENTITY}_1, \underline{\texttt{Relation}}, \texttt{ENTITY}_2)$

\paragraph{\emph{\textbf{Cleaning.}}} To improve the quality of the retrieved triplets in $\mathcal{F}_1$, we further employ a closed-source LLM (GPT-4 \cite{achiam2023gpt}) via the OpenAI API, denoted as $LLM_2$. This model curates a subset of triplets, $\mathcal{F}^\prime_1 \subset \mathcal{F}_1$, by assessing their validity. While $\mathcal{F}_1$ is designed to be comprehensive, ensuring that no meaningful relations are overlooked, $\mathcal{F}^\prime_1$ prioritizes accuracy, filtering out false or misleading relations to maintain the integrity of the constructed KG. The subset curation process can be expressed as:
 
$\text{Prompt}_2(\texttt{ENTITY}_1, \underline{\texttt{Relation}}, \texttt{ENTITY}_2)\xrightarrow{LLM_2} \emph{\texttt{True}}/\emph{\texttt{False}}$\\
We take the triplet subset $\mathcal{F}^\prime_1$ and the associated labels from $LLM_2$ to train a classifier $f_1$ to clean the remaining triplets $\mathcal{F}_1 \setminus \mathcal{F}^\prime_1$. This identifies the plausible triplets in a time- and cost-effective manner, producing the cleaned triplet set $\mathcal{F}_2$.

\paragraph{\emph{\textbf{Refinement.}}} The triplets retrieved and filtered through LLMs often contain lexically similar but distinct relations, such as \texttt{is treated with} and \texttt{is treated using}, which should be unified for effective representation learning in the medical KG. Following \cite{jiang2023graphcare}, we first extract word embeddings for all relations using a pre-trained language model (BERT \cite{kenton2019bert}) and apply agglomerative clustering \cite{rokach2005clustering} to group lexically similar relations. However, lexical similarity does not always equate to semantic similarity—relations with opposing meanings may differ by only a few words, e.g., \texttt{be not typically associated with} and \texttt{be often associated with}. To address this, we leverage an open-source LLM to detect and refine clusters containing semantically contradictory relations. Finally, we obtain the high-quality triplet set $\mathcal{F}$ for medical KG. 

The three-step process ensures the construction of a more expressive and robust medical KG, which serves as the foundation for subsequent model stages. Further details on the reliability of the generated KG are provided in Appendix~\ref{sec:appendix_c}.

\subsection{Hierarchical Representation Learning}
Building on the medical KG, we design a hierarchical representation learning process, which can be seen in Figure~\ref{fig:architecture}b. To obtain \textbf{code-level representations}, we use the medical KG $\mathcal{G}=(\mathcal{C}, \mathcal{R}, \mathcal{F})$ along with patient data $\mathcal{V}_i$ as input to the local encoder at the code level. Specifically, we employ a multi-relational graph convolutional network, CompGCN \citep{vashishth2020composition}, to iteratively update the attributes of entities and relations with global medical knowledge. The representations are updated as follows:
\begin{equation}
\vc_v^{l} = \sigma\left(\sum_{(u,r_i) \in \mathcal{N}_v} \mW^l_{\textnormal{ent}}\phi(\vc_{u}^{l-1}, \vr_i^{l-1})\right), \quad \vr_i^l = \mW^l_{\textnormal{rel}}\vr_i^{l-1},
\label{eq:gcn_update}
\end{equation}
where $\vc^l_v$ and $\vr^l_i$ denote the layer-$l$ representations of entity $v$ and relation $i$, respectively. $\mW_{\text{ent}}^l$ and $\mW_{\text{rel}}^l$ are the learnable weight matrices for updating entities and relations at layer $l$, while $\sigma$ denotes an activation function. The set $\mathcal{N}_v$ consists of the neighbors of entity $v$ connected via the associated relations $r_i$. The function $\phi$ represents the circular correlation. 

After global message passing, the updated representations of medical codes are processed by the code-level prototype-based encoder, $PEnc^{c}(\cdot)$, which incorporates shared information across codes through prototype learning. The details of this prototype-based encoder are provided in Section~\ref{sec:proto_enc}. Following this step, we obtain the updated code representation, denoted as $\vc_{i,j,k} \in \mathbb{R}^{d}$, where $d$ is the embedding dimension, for the $k$-th medical code recorded in the $j$-th visit of the $i$-th patient.

The \textbf{visit-level representations} are obtained by passing the code representations into the visit-level local encoder. Here we simply use average pooling, and the representation for $j$-th visit of $i$-th patient is calculated based on the representations of medical codes recorded in this visit:
\begin{equation}
    \vv^\prime_{i,j} = \frac{1}{|\mathcal{V}_{i,j}|}\sum_{k=1}^{|\mathcal{V}_{i,j}|} \vc_{i,j,k}.
\end{equation}
Similarly, $\vv^\prime_{i,j}$ is fed into the visit-level prototype-based encoder $PEnc^v(\cdot)$ with the updated visit representation $\vv_{i,j}$ obtained.

To obtain the \textbf{patient-level representations}, we account for the temporal sequence of multiple visits and employ a Transformer-based encoder to capture interactions among them. Given that recent visits have a stronger impact on future healthcare predictions, we use the representation of the last visit after the Transformer encoder as the patient-level representation $\vp_i'$. This representation is further enhanced through the patient-level prototype-based encoder, $PEnc^p(\cdot)$. This enables us to obtain the updated patient representation $\vp_i$.

\subsection{Prototype-based Encoder}
\label{sec:proto_enc}
We now provide a detailed explanation of the prototype-based encoder $PEnc(\cdot)$, which is applied at all three levels of the hierarchy. This module (Figure~\ref{fig:architecture}c) consists of two key steps: prototype learning and prototype infusion. 

\textbf{Prototype Learning.} The first step takes the representations of objects (codes, visits, or patients) within the same level and learnable prototypes as input. For each level of the hierarchy, the prototypes are randomly initialized and the number of prototypes is a hyperparameter. Without loss of generality, let the input representations be denoted as $\mX^\prime\in \mathbb{R}^{n \times d}$ and the learnable prototypes as $\mH \in \mathbb{R}^{m \times d}$, where $n$ is the number of objects, $m$ is the number of prototypes, and $d$ is the embedding dimension. To enable prototypes to absorb intrinsic similarities from the objects, we employ a cross-attention mechanism, formulated as the equation below:
\begin{equation}
    \hat{\mH} = \text{Softmax}(\frac{(\mH \mW_Q)(\mX^\prime \mW_K)^\top}{\sqrt{d}})\mX^\prime \mW_V,
\end{equation}
where $\mW_Q$, $\mW_K$, and $\mW_V$ are learnable parameters. This mechanism enhances the prototypes by integrating information from the objects at the corresponding level, enabling a more structured and meaningful representation.

\textbf{Prototype Infusion.} With the enhanced prototypes obtained in the previous step, the second step infuses the intrinsic similarities captured by the prototypes into the object representations. For each input object representation $\vx^\prime_i$, we apply the similarity-weighted summation idea for infusion \cite{liang2024clusterfomer}, formulated as follows:
\begin{equation}
   \vx_i = \vx^\prime_i + \frac{1}{m}\sum_{j=1}^m \alpha_{i,j}\mW_I\hat{\vh}_j,\ \text{with}\ \alpha_{i,j} = \text{Softmax}(\frac{{\vx^\prime_i}^\top \hat{\vh}_j}{\|\vx^\prime_i\| \|\hat{\vh}_j\|}),
\end{equation}
where $\mW_I$ represents learnable parameters, and $\alpha_{i,j}$ denotes the normalized similarity between object $\vx_i$ and enhanced prototype $\hat{\vh}_j$. The resulting prototype-infused object representations, $\mX$, are subsequently used to compute representations at the next level in the code-visit-patient hierarchy.

\subsection{Hierarchical Fusion}
\label{sec:hf}
We now detail the hierarchical fusion module used to generate the final patient representation. The structure of this module is outlined in Figure~\ref{fig:architecture}d. Specifically, for each patient $i$, the patient-level representation $\vp_i$ is fused with prototypes from all three hierarchical levels—code, visit, and patient—denoted as $\mH^{c}$, $\mH^{v}$, and $\mH^{p}$, respectively. This fusion is achieved using a cross-attention mechanism, formulated as follows:
\begin{equation}
    \vp^t_i = \text{Softmax}(\frac{(\vp_i\mW_Q^t)({\mH^t}\mW_K^t)^\top}{\sqrt{d}})\mH^t {\mW_V^t},\ t \in \{c, v, p\}
\end{equation}
The resulting level-specific representations $\{\vp^t_i|t\in\{c,v,p\}\}$ are then combined to form the final patient representation:
\begin{equation}
    \vp_i^{\text{final}} =\!\! \sum_{t\in\{c,v,p\}}\!\! \!\!\beta_t \vp^t_i,\ \text{with}\ \beta_t = \frac{\exp({\vp^t_i}^\top\vw_F/\tau)}{\sum_{t\in\{c,v,p\}}\exp({\vp^t_i}^\top\vw_F/\tau)}.
\end{equation}
Here, $\tau$ is the softmax temperature, and $\beta_t$ represents the contribution weight of each level, computed based on the normalized similarity between the learnable vector $\vw_F \in \mathbb{R}^d$ and $\vp_i^t$, thereby determining the relative importance of information from each hierarchical level to the prediction, enabling a more interpretable and effective modeling approach.

For healthcare prediction, the final patient representation is passed through a linear projection layer to generate the predicted outcome, i.e., $\hat y_i = \text{Linear}(\vp_i^{\text{final}})$. The whole framework is trained with a task-specific loss, which depends on the prediction task \cite{jiang2023graphcare}: binary cross-entropy loss is used for binary classification and multi-label classification tasks, while cross-entropy loss is applied for multi-class classification tasks.

\begin{table}[htbp]
\caption{Basic information about MIMIC-III and MIMIC-IV after preprocessing. c, v, and p are the abbreviations for code, visit, and patient respectively. \#v/p and \#c/p separately denotes the average number of visits per patient and the average number of codes per visit.}
\vspace{-10px}
\label{tab:dataset_statistics}
\centering
\begin{tabular}{cccccc}
\toprule
\textbf{Dataset} & \textbf{\#Patients} & \textbf{\#Visits} & \textbf{\#Codes} & \textbf{\#v/p} & \textbf{\#c/v} \\
\toprule
MIMIC-III  & 5,453 & 14,330 & 657 & 2.63 & 39.7\\
MIMIC-IV  & 51,473 & 167,042 & 708 & 3.25 & 21.6\\
\bottomrule
\end{tabular}
\vspace{-0.45cm}
\end{table}

\section{Experiments}
\subsection{Experimental Setup}

\paragraph{\emph{\textbf{Datasets.}}} We use two real-world medical datasets, MIMIC-III \citep{johnson2016mimic} and MIMIC-IV \citep{johnson2020mimic} for experiments. The basic statistics of the datasets are presented in Table~\ref{tab:dataset_statistics}. For MIMIC-III, a sliding window is employed to augment the sample size for all tasks other than mortality prediction. We split both datasets into train/validation/test sets with a ratio of 6:2:2.

\begin{table*}[t!]
  \centering
  \caption{Results comparison for five tasks on MIMIC-III and MIMIC-IV. Best
    results are in bold and the second best results are underlined. Performances are reported in the form of mean(std). ProtoEHR achieves best or second-best performance on {\color{black}24/24} metrics across all five tasks on two datasets, demonstrating robust generalization across datasets and clinical prediction targets.}
  \label{tab:main_exp}
\vspace{-10px}
  %----- Subtable for Tasks 1–3 -----
  \begin{subtable}{\textwidth}
    \centering
    \setlength{\tabcolsep}{4pt}
    \begin{tabular}{%
      l
      |cc|cc   % Task 1
      |cc|cc   % Task 2
      |cc|cc   % Task 3
    }
      \toprule
      & \multicolumn{4}{c|}{\textbf{Task 1: Mortality Prediction}}
      & \multicolumn{4}{c|}{\textbf{Task 2: Readmission Prediction}}
      & \multicolumn{4}{c}{\textbf{Task 3: Length of Stay Prediction}}
      \\
      \textbf{Model}
        & \multicolumn{2}{c|}{\textbf{MIMIC-III}}
        & \multicolumn{2}{c|}{\textbf{MIMIC-IV}}
        & \multicolumn{2}{c|}{\textbf{MIMIC-III}}
        & \multicolumn{2}{c|}{\textbf{MIMIC-IV}}
        & \multicolumn{2}{c|}{\textbf{MIMIC-III}}
        & \multicolumn{2}{c}{\textbf{MIMIC-IV}}
      \\
      \cmidrule(lr){2-3}\cmidrule(lr){4-5}
      \cmidrule(lr){6-7}\cmidrule(lr){8-9}
      \cmidrule(lr){10-11}\cmidrule(lr){12-13}
        & AUPRC & AUROC & AUPRC & AUROC
        & AUPRC & AUROC & AUPRC & AUROC
        & AUROC  & F1   & AUROC  & F1
      \\
      \midrule
      Deepr
& $\textnormal{8.2}_{\pm\textnormal{2.4}}$ & $\textnormal{67.4}_{\pm\textnormal{4.4}}$
& $\underline{14.5}_{\pm\textnormal{2.9}}$ & $\textnormal{88.2}_{\pm\textnormal{1.6}}$
& $\textnormal{30.9}_{\pm\textnormal{2.1}}$ & $\textnormal{59.1}_{\pm\textnormal{2.0}}$
& $\textnormal{54.3}_{\pm\textnormal{0.9}}$ & $\underline{70.6}_{\pm\textnormal{0.5}}$
& $\textnormal{72.3}_{\pm\textnormal{0.9}}$ & $\textnormal{22.2}_{\pm\textnormal{1.2}}$
& $\textnormal{80.8}_{\pm\textnormal{0.2}}$ & $\textnormal{28.0}_{\pm\textnormal{0.5}}$ \\

AdaCare
& $\textnormal{6.0}_{\pm\textnormal{1.6}}$ & $\textnormal{61.5}_{\pm\textnormal{5.5}}$
& $\textnormal{12.7}_{\pm\textnormal{2.5}}$ & $\textnormal{88.4}_{\pm\textnormal{1.5}}$
& $\textnormal{31.4}_{\pm\textnormal{2.2}}$ & $\underline{60.1}_{\pm\textnormal{2.0}}$
& $\underline{54.4}_{\pm\textnormal{0.9}}$ & $\underline{70.6}_{\pm\textnormal{0.5}}$
& $\underline{73.0}_{\pm\textnormal{0.8}}$ & $\underline{22.7}_{\pm\textnormal{1.2}}$
& $\underline{82.4}_{\pm\textnormal{0.2}}$ & $\underline{29.7}_{\pm\textnormal{0.4}}$ \\

GRASP
& $\underline{9.2}_{\pm\textnormal{3.1}}$ & $\underline{68.5}_{\pm\textnormal{4.2}}$
& $\textnormal{13.6}_{\pm\textnormal{2.7}}$ & $\underline{88.8}_{\pm\textnormal{1.4}}$
& $\textnormal{29.5}_{\pm\textnormal{2.1}}$ & $\textnormal{57.0}_{\pm\textnormal{2.1}}$
& $\textnormal{53.0}_{\pm\textnormal{0.9}}$ & $\textnormal{69.6}_{\pm\textnormal{0.6}}$
& $\textnormal{71.2}_{\pm\textnormal{0.9}}$ & $\textnormal{18.1}_{\pm\textnormal{1.0}}$
& $\textnormal{81.1}_{\pm\textnormal{0.2}}$ & $\textnormal{26.0}_{\pm\textnormal{0.4}}$ \\

StageNet
& $\textnormal{5.5}_{\pm\textnormal{1.8}}$ & $\textnormal{59.4}_{\pm\textnormal{4.5}}$
& $\textnormal{13.2}_{\pm\textnormal{2.9}}$ & $\textnormal{87.3}_{\pm\textnormal{1.7}}$
& $\textbf{35.2}_{\pm\textnormal{2.7}}$ & $\textnormal{60.0}_{\pm\textnormal{2.0}}$
& $\textnormal{50.2}_{\pm\textnormal{1.0}}$ & $\textnormal{68.7}_{\pm\textnormal{0.6}}$
& $\textnormal{71.9}_{\pm\textnormal{0.9}}$ & $\textnormal{22.3}_{\pm\textnormal{1.1}}$
& $\textnormal{81.3}_{\pm\textnormal{0.2}}$ & $\textnormal{27.4}_{\pm\textnormal{0.5}}$ \\

GraphCare
& $\textnormal{5.6}_{\pm\textnormal{1.5}}$ & $\textnormal{58.4}_{\pm\textnormal{4.3}}$
& $\textnormal{9.2}_{\pm\textnormal{2.2}}$ & $\textnormal{80.8}_{\pm\textnormal{1.9}}$
& $\textnormal{31.7}_{\pm\textnormal{2.5}}$ & $\textnormal{58.5}_{\pm\textnormal{2.2}}$
& $\textnormal{52.3}_{\pm\textnormal{0.9}}$ & $\textnormal{69.1}_{\pm\textnormal{0.6}}$
& $\textnormal{65.0}_{\pm\textnormal{0.8}}$ & $\textnormal{13.4}_{\pm\textnormal{0.7}}$
& $\textnormal{69.2}_{\pm\textnormal{0.4}}$ & $\textnormal{19.6}_{\pm\textnormal{0.4}}$ \\
KerPrint
& $\textnormal{5.0}_{\pm\textnormal{1.3}}$ & $\textnormal{58.8}_{\pm\textnormal{4.7}}$
& $\textnormal{10.6}_{\pm\textnormal{2.5}}$ & $\textnormal{83.7}_{\pm\textnormal{2.2}}$
& $\textnormal{29.4}_{\pm\textnormal{2.0}}$ & $\textnormal{57.8}_{\pm\textnormal{1.9}}$
& $\textnormal{53.4}_{\pm\textnormal{0.9}}$ & $\textnormal{70.2}_{\pm\textnormal{0.5}}$
& $\textnormal{70.9}_{\pm\textnormal{0.9}}$ & $\textnormal{21.6}_{\pm\textnormal{1.2}}$
& $\textnormal{77.9}_{\pm\textnormal{0.3}}$ & $\textnormal{26.7}_{\pm\textnormal{0.4}}$ \\
\midrule\midrule
ProtoEHR
& $\textbf{11.1}_{\pm\textnormal{3.7}}$ & $\textbf{71.4}_{\pm\textnormal{3.1}}$
& $\textbf{17.0}_{\pm\textnormal{3.3}}$ & $\textbf{89.2}_{\pm\textnormal{1.3}}$
& $\underline{33.0}_{\pm\textnormal{2.5}}$ & $\textbf{61.7}_{\pm\textnormal{2.2}}$
& $\textbf{54.5}_{\pm\textnormal{0.9}}$ & $\textbf{70.9}_{\pm\textnormal{0.5}}$
& $\textbf{75.3}_{\pm\textnormal{0.9}}$ & $\textbf{23.6}_{\pm\textnormal{1.2}}$
& $\textbf{83.2}_{\pm\textnormal{0.2}}$ & $\textbf{31.9}_{\pm\textnormal{0.5}}$ \\

      \bottomrule
    \end{tabular}
  \end{subtable}

  % \bigskip % extra vertical space

  %----- Subtable for Tasks 4–5 -----
  \begin{subtable}{\textwidth}
    \centering
    \setlength{\tabcolsep}{4pt}
    \begin{tabular}{%
      l
      |ccc|ccc   % Task 4
      |ccc|ccc   % Task 5
    }
      \toprule
      & \multicolumn{6}{c|}{\textbf{Task 4: Drug Recommendation}}
      & \multicolumn{6}{c}{\textbf{Task 5: Phenotype Prediction}}
      \\
      \textbf{Model}
        & \multicolumn{3}{c|}{\textbf{MIMIC-III}}
        & \multicolumn{3}{c|}{\textbf{MIMIC-IV}}
        & \multicolumn{3}{c|}{\textbf{MIMIC-III}}
        & \multicolumn{3}{c}{\textbf{MIMIC-IV}}
      \\
      \cmidrule(lr){2-4}\cmidrule(lr){5-7}
      \cmidrule(lr){8-10}\cmidrule(lr){11-13}
        & AUPRC & AUROC & F1    & AUPRC & AUROC & F1
        & AUPRC & AUROC & F1    & AUPRC & AUROC & F1
      \\
      \midrule
Deepr
& $\textnormal{67.0}_{\pm\textnormal{0.6}}$ & $\underline{91.0}_{\pm\textnormal{0.3}}$ & $\textnormal{50.9}_{\pm\textnormal{0.7}}$
& $\textnormal{70.6}_{\pm\textnormal{0.2}}$ & $\textnormal{95.6}_{\pm\textnormal{0.0}}$ & $\textnormal{56.6}_{\pm\textnormal{0.3}}$
& $\textnormal{63.4}_{\pm\textnormal{0.8}}$ & $\textnormal{83.6}_{\pm\textnormal{0.4}}$ & $\textnormal{42.9}_{\pm\textnormal{0.9}}$
& $\textnormal{77.5}_{\pm\textnormal{0.3}}$ & $\textnormal{92.6}_{\pm\textnormal{0.1}}$ & $\textnormal{60.9}_{\pm\textnormal{0.4}}$ \\

AdaCare
& $\textnormal{62.6}_{\pm\textnormal{0.7}}$ & $\textnormal{88.0}_{\pm\textnormal{0.4}}$ & $\textnormal{44.5}_{\pm\textnormal{0.5}}$
& $\textnormal{69.3}_{\pm\textnormal{0.2}}$ & $\textnormal{94.7}_{\pm\textnormal{0.1}}$ & $\textnormal{56.0}_{\pm\textnormal{0.3}}$
& $\textnormal{57.6}_{\pm\textnormal{0.8}}$ & $\textnormal{80.6}_{\pm\textnormal{0.4}}$ & $\textnormal{31.5}_{\pm\textnormal{0.9}}$
& $\textnormal{75.9}_{\pm\textnormal{0.3}}$ & $\textnormal{91.9}_{\pm\textnormal{0.1}}$ & $\textnormal{59.9}_{\pm\textnormal{0.4}}$ \\

GRASP
& $\textnormal{63.0}_{\pm\textnormal{0.7}}$ & $\textnormal{89.7}_{\pm\textnormal{0.3}}$ & $\textnormal{45.8}_{\pm\textnormal{0.5}}$
& $\textnormal{68.9}_{\pm\textnormal{0.2}}$ & $\textnormal{95.2}_{\pm\textnormal{0.0}}$ & $\textnormal{55.4}_{\pm\textnormal{0.3}}$
& $\textnormal{57.6}_{\pm\textnormal{0.7}}$ & $\textnormal{80.1}_{\pm\textnormal{0.5}}$ & $\textnormal{31.9}_{\pm\textnormal{0.9}}$
& $\textnormal{76.2}_{\pm\textnormal{0.3}}$ & $\textnormal{91.9}_{\pm\textnormal{0.1}}$ & $\textnormal{60.7}_{\pm\textnormal{0.4}}$ \\

StageNet
& $\textnormal{66.0}_{\pm\textnormal{0.6}}$ & $\textnormal{90.4}_{\pm\textnormal{0.3}}$ & $\textnormal{50.7}_{\pm\textnormal{0.7}}$
& $\textnormal{70.5}_{\pm\textnormal{0.2}}$ & $\textnormal{95.4}_{\pm\textnormal{0.0}}$ & $\textnormal{57.4}_{\pm\textnormal{0.3}}$
& $\textnormal{62.6}_{\pm\textnormal{0.7}}$ & $\textnormal{83.1}_{\pm\textnormal{0.4}}$ & $\textnormal{42.3}_{\pm\textnormal{0.9}}$
& $\textnormal{77.5}_{\pm\textnormal{0.3}}$ & $\textnormal{92.6}_{\pm\textnormal{0.1}}$ & $\textnormal{60.4}_{\pm\textnormal{0.4}}$ \\

GraphCare
& $\textnormal{65.9}_{\pm\textnormal{0.6}}$ & $\textnormal{84.3}_{\pm\textnormal{0.4}}$ & $\textnormal{43.7}_{\pm\textnormal{0.9}}$
& $\textnormal{64.7}_{\pm\textnormal{0.2}}$ & $\textnormal{94.4}_{\pm\textnormal{0.1}}$ & $\textnormal{46.6}_{\pm\textnormal{0.2}}$
& $\textnormal{62.4}_{\pm\textnormal{0.6}}$ & $\textnormal{82.5}_{\pm\textnormal{0.4}}$ & $\textnormal{43.8}_{\pm\textnormal{0.7}}$
& $\textnormal{73.0}_{\pm\textnormal{0.3}}$ & $\textnormal{90.0}_{\pm\textnormal{0.1}}$ & $\textnormal{51.3}_{\pm\textnormal{0.4}}$ \\

KerPrint
& $\underline{67.9}_{\pm\textnormal{0.7}}$ & $\underline{91.0}_{\pm\textnormal{0.3}}$ & $\underline{52.3}_{\pm\textnormal{0.7}}$
& $\underline{72.1}_{\pm\textnormal{0.2}}$ & $\textbf{95.9}_{\pm\textnormal{0.0}}$ & $\underline{58.7}_{\pm\textnormal{0.2}}$
& $\underline{67.8}_{\pm\textnormal{0.8}}$ & $\underline{85.4}_{\pm\textnormal{0.4}}$ & $\textbf{50.2}_{\pm\textnormal{1.0}}$
& $\textbf{78.7}_{\pm\textnormal{0.3}}$ & $\textbf{93.3}_{\pm\textnormal{0.1}}$ & $\underline{63.6}_{\pm\textnormal{0.4}}$ \\
\midrule\midrule
ProtoEHR
& $\textbf{70.6}_{\pm\textnormal{0.6}}$ & $\textbf{91.7}_{\pm\textnormal{0.3}}$ & $\textbf{54.4}_{\pm\textnormal{0.8}}$
& $\textbf{72.4}_{\pm\textnormal{0.2}}$ & $\underline{95.8}_{\pm\textnormal{0.1}}$ & $\textbf{60.0}_{\pm\textnormal{0.2}}$
& $\textbf{68.2}_{\pm\textnormal{0.7}}$ & $\textbf{86.1}_{\pm\textnormal{0.4}}$ & $\underline{45.6}_{\pm\textnormal{1.0}}$
& $\underline{78.5}_{\pm\textnormal{0.3}}$ & $\underline{93.2}_{\pm\textnormal{0.1}}$ & $\textbf{64.2}_{\pm\textnormal{0.4}}$ \\

      \bottomrule
    \end{tabular}
  \end{subtable}

\end{table*}

\textbf{Tasks and Metrics.} Five tasks were used for evaluation: 
\begin{itemize}[leftmargin=10px]
    \item \textbf{Mortality Prediction.} An imbalanced binary classification task with label $y_i \in \{0,1\}$ where $y_i = 1$ indicates mortality. For a patient with $|\mathcal{V}_i|$ visits, the model takes data $(\mathcal{V}_{i,1},\ \dots,\ \mathcal{V}_{i, |\mathcal{V}_i|-1})$ to predict if the patient is deceased in the visit $\mathcal{V}_{i,|\mathcal{V}_i|}$ and if it occurs within 30 days from the visit $\mathcal{V}_{i, |\mathcal{V}_i|-1}$.
    \item \textbf{Readmission Prediction.} A binary classification task with label $y_i \in \{0,1\}$ where $y_i = 1$ indicates readmission. For a patient with $|\mathcal{V}_i|$ visits, the model takes data $(\mathcal{V}_{i,1},\ \dots,\ \mathcal{V}_{i, |\mathcal{V}_i|-1})$ to predict if visit $\mathcal{V}_{i,|\mathcal{V}_i|}$ occurs within 30 days from the visit $\mathcal{V}_{i, |\mathcal{V}_i|- 1}$.
    \item \textbf{Length-of-stay Prediction} A multi-class classification task with 10 labels indicating the duration of the patient's hospital visit $y_i \in \{0,\ 1,\ \dots,\ 9\}$. For a patient with $|\mathcal{V}_i|$ visits, the model takes data $(\mathcal{V}_{i,1},\ \dots,\ \mathcal{V}_{i, |\mathcal{V}_i|})$ to predict the duration of the visit $\mathcal{V}_{i,|\mathcal{V}_i|}$. The label $y_i = 0$ represents discharge within one day, the labels $y_i \in \{1,\ \dots,\ 7\}$ indicate that the duration of stay is $\{y_i \leq t < y_i+1\}_{y_i=1}^7$ days, the label $y_i = 8$ means the patient's stay is $7 < t \leq 14$ days, and the label $y_i = 9$ indicates that the stay is longer than 14 days.
    \item \textbf{Drug Recommendation} A multi-label classification task where the model takes data $(\mathcal{V}_{i,1},\ \dots,\ \mathcal{V}_{i, |\mathcal{V}_i| - 1})\ \cup\ \mathcal{V}'_{i, |\mathcal{V}_i|}$ to predict the prescriptions on visit $\mathcal{V}_{i, |\mathcal{V}_i|}$. Notation $\mathcal{V}'_{i, |\mathcal{V}_i|}$ denotes the data of the last visit excluding prescription codes. The output is a multi-hot vector $\hat \vy_i \in \mathbb{R}^{202}$ where the set of prescriptions contain 201 elements, the last dimension is used to indicate no prescriptions on the last visit.
    \item \textbf{Phenotype Prediction} A multi-label classification tasks where the model takes data $(\mathcal{V}_{i,1},\ \dots,\ \mathcal{V}_{i, |\mathcal{V}_i| - 1})\ \cup\ \mathcal{V}''_{i, |\mathcal{V}_i|}$ to predict the phenotypes on visit $\mathcal{V}_{i, |\mathcal{V}_i|}$. Notation $\mathcal{V}''_{i, |\mathcal{V}_i|}$ denotes the data of the last visit excluding diagnosis codes. The output is a multi-hot vector $\hat \vy_i \in \mathbb{R}^{26}$ where the set of phenotypes contains 25 diagnoses \cite{harutyunyan2019multitask}, and the last dimension is used to indicate no phenotype is present on the last visit.
\end{itemize}
Three different metrics were used across tasks for performance evaluation. The area under the receiver operating characteristic curve (\textbf{AUROC}) quantifies the model's ability to differentiate between classes; the area under the precision-recall curve (\textbf{AUPRC}) is similar to AUROC but emphasizes measuring performance on separating imbalanced classes; the \textbf{F1} score is a balance between precision and recall, capturing the importance of false positives and false negatives. 
For these metrics, we report both the performance mean and the standard deviation (std) of bootstrapping 100 times.

\paragraph{\emph{\textbf{Baselines.}}}  We compare the performance for different tasks against several representative baselines: Deepr \cite{nguyen2016deepr}, AdaCare \cite{ma2020adacare}, GRASP \cite{zhang2021grasp}, StageNet \cite{gao2020stagenet}, GraphCare \cite{jiang2023graphcare}, and KerPrint \cite{yang2023kerprint}. For each baseline, we perform hyperparameter tuning based on the recommended search ranges specified in their original papers to ensure a fair and competitive comparison. For methods that incorporate a KG in their modeling, we use the same medical KG constructed in our framework to ensure consistency across evaluations.

\paragraph{\emph{\textbf{Implementation.}}} In our implementation, the Adam gradient optimizer and ExponentialLR scheduler are used. To optimize model performance, hyperparameter search is performed using grid search. It should also be noted that this can be implemented more efficiently with Bayesian-based search methods such as Optuna \cite{akiba2019optuna}. The parameters considered include CompGCN layers \{1,2,3,4\}, transformer encoder depth \{1,2,4\}, number of code-level prototypes \{32,64\}, number of visit-level prototypes \{4,8,16,32\}, number of patient-level prototypes \{2,4,8,16\}, dropout probability \{0.1, 0.3, 0.5\}, and learning rate \{0.0001, 0.0005, 0.001\}. We use early stopping to prevent overfitting for all models, and AUPRC on the validation set is used as the early stopping metric for all tasks apart from the length-of-stay prediction, for which we use AUROC. The early stopping threshold is set to 20 epochs. The implementation codes are available here\footnote{\url{https://github.com/caizicharles/ProtoEHR.git}}.

\subsection{Performance Evaluation}
Table~\ref{tab:main_exp} displays the results for ProtoEHR and the baselines on both datasets across five tasks. Our approach outperforms all baselines on almost all of the tasks, demonstrating the effectiveness of modeling within and across hierarchical levels for prediction. We can observe that on MIMIC-III, ProtoEHR outperforms the baselines most noticeably in predicting mortality, achieving a 20.7\% AUPRC improvement. On MIMIC-IV, our approach achieves the best results in predicting mortality and length of stay, achieving a 17.2\% AUPRC improvement and a 7.4\% F1-score improvement, respectively. Since mortality prediction is a patient-level task (as explained in Section~\ref{sec:hf_interp}), achieving superior performance on this task shows that explicitly modeling the hierarchical structure and incorporating patient-level similarity are crucial in helping the model distinguish patients of imbalanced classes.

Among the baselines, GRASP is better at predicting mortality due to its use of prototypes at the patient's level, further demonstrating the importance of capturing intrinsic similarity in the EHR data. In comparison, KerPrint achieves competitive performance on drug recommendation and phenotype prediction because it models the complex relations within the code level and integrates it directly with the learned patient representation, enabling the model to apply code-level knowledge to code-level tasks. Regarding GraphCare, its original implementation relies on GPT-4 to retrieve all possible knowledge triplets associated with each entity, resulting in a highly enriched knowledge graph that includes many entities not present in the original EHR dataset. This external augmentation introduces additional medical knowledge but is computationally expensive and not feasible in our setting. To ensure fairness and reproducibility, we use GPT-4 to generate a smaller set of triplets, limiting the entities to medical codes only. The relatively lower performance of GraphCare under our KG construction suggests that its effectiveness is heavily dependent on the richness of external knowledge. In contrast, our proposed ProtoEHR achieves robust performance by fully leveraging both within-level similarities and cross-level hierarchical structures in the EHR data, demonstrating effectiveness even with a lightweight and clinically grounded KG.

\begin{table*}[t!]
\centering
\caption{Ablation study of our proposed method for five tasks on MIMIC-IV. Notations w/o, Proto., and HF abbreviates without, prototype, and hierarchical fusion respectively.}
\vspace{-10px}
\label{tab:abl}
\setlength{\tabcolsep}{2pt}
\begin{tabular}{l
   cc  cc  cc   ccc  ccc}
\toprule
& \multicolumn{2}{c}{\textbf{Task 1: Mortality}}
& \multicolumn{2}{c}{\textbf{Task 2: Readmission}}
& \multicolumn{2}{c}{\textbf{Task 3: LOS}}
& \multicolumn{3}{c}{\textbf{Task 4: Drug Rec.}}
& \multicolumn{3}{c}{\textbf{Task 5: Phenotype}}
\\
\cmidrule(lr){2-3} \cmidrule(lr){4-5} \cmidrule(lr){6-7}
\cmidrule(lr){8-10} \cmidrule(lr){11-13}
\textbf{Model}
& AUPRC & AUROC
& AUPRC & AUROC
& AUROC  & F1
& AUPRC & AUROC & F1
& AUPRC & AUROC & F1
\\

\midrule
w/ ALL
& $\textnormal{17.0}_{\pm\textnormal{3.3}}$ & $\textnormal{89.2}_{\pm\textnormal{1.3}}$
& $\textnormal{54.5}_{\pm\textnormal{0.9}}$ & $\textnormal{70.9}_{\pm\textnormal{0.5}}$
& $\textnormal{83.2}_{\pm\textnormal{0.2}}$ & $\textnormal{31.9}_{\pm\textnormal{0.5}}$
& $\textnormal{72.4}_{\pm\textnormal{0.2}}$ & $\textnormal{95.8}_{\pm\textnormal{0.1}}$ & $\textnormal{60.0}_{\pm\textnormal{0.2}}$
& $\textnormal{78.5}_{\pm\textnormal{0.3}}$ & $\textnormal{93.2}_{\pm\textnormal{0.1}}$ & $\textnormal{64.2}_{\pm\textnormal{0.4}}$
\\
\midrule

w/o Medical KG
& $\textnormal{8.9}_{\pm\textnormal{1.5}}$  & $\textnormal{86.7}_{\pm\textnormal{1.8}}$
& $\textnormal{52.1}_{\pm\textnormal{0.9}}$ & $\textnormal{68.8}_{\pm\textnormal{0.6}}$
& $\textnormal{81.6}_{\pm\textnormal{0.2}}$ & $\textnormal{29.2}_{\pm\textnormal{0.4}}$
& $\textnormal{71.4}_{\pm\textnormal{0.2}}$ & $\textnormal{95.4}_{\pm\textnormal{0.1}}$ & $\textnormal{59.3}_{\pm\textnormal{0.3}}$
& $\textnormal{77.4}_{\pm\textnormal{0.3}}$ & $\textnormal{92.6}_{\pm\textnormal{0.1}}$ & $\textnormal{61.7}_{\pm\textnormal{0.4}}$
\\

w/o Code Proto.
& $\textnormal{16.1}_{\pm\textnormal{3.6}}$ & $\textnormal{88.8}_{\pm\textnormal{1.5}}$
& $\textnormal{52.6}_{\pm\textnormal{0.9}}$ & $\textnormal{69.9}_{\pm\textnormal{0.5}}$
& $\textnormal{82.5}_{\pm\textnormal{0.2}}$ & $\textnormal{30.2}_{\pm\textnormal{0.5}}$
& $\textnormal{71.7}_{\pm\textnormal{0.2}}$ & $\textnormal{95.5}_{\pm\textnormal{0.1}}$ & $\textnormal{60.0}_{\pm\textnormal{0.2}}$
& $\textnormal{78.4}_{\pm\textnormal{0.3}}$ & $\textnormal{93.2}_{\pm\textnormal{0.1}}$ & $\textnormal{63.3}_{\pm\textnormal{0.3}}$
\\

w/o Visit Proto.
& $\textnormal{13.5}_{\pm\textnormal{2.7}}$ & $\textnormal{86.8}_{\pm\textnormal{1.6}}$
& $\textnormal{52.8}_{\pm\textnormal{0.9}}$ & $\textnormal{69.3}_{\pm\textnormal{0.6}}$
& $\textnormal{82.8}_{\pm\textnormal{0.2}}$ & $\textnormal{30.3}_{\pm\textnormal{0.5}}$
& $\textnormal{71.9}_{\pm\textnormal{0.2}}$ & $\textnormal{95.6}_{\pm\textnormal{0.1}}$ & $\textnormal{59.4}_{\pm\textnormal{0.2}}$
& $\textnormal{78.3}_{\pm\textnormal{0.3}}$ & $\textnormal{93.1}_{\pm\textnormal{0.1}}$ & $\textnormal{63.6}_{\pm\textnormal{0.4}}$
\\

w/o Patient Proto.
& $\textnormal{13.6}_{\pm\textnormal{2.8}}$ & $\textnormal{86.3}_{\pm\textnormal{1.7}}$
& $\textnormal{53.3}_{\pm\textnormal{0.9}}$ & $\textnormal{70.2}_{\pm\textnormal{0.5}}$
& $\textnormal{83.0}_{\pm\textnormal{0.2}}$ & $\textnormal{30.6}_{\pm\textnormal{0.5}}$
& $\textnormal{71.3}_{\pm\textnormal{0.2}}$ & $\textnormal{95.5}_{\pm\textnormal{0.1}}$ & $\textnormal{59.8}_{\pm\textnormal{0.2}}$
& $\textnormal{78.1}_{\pm\textnormal{0.3}}$ & $\textnormal{93.0}_{\pm\textnormal{0.1}}$ & $\textnormal{63.6}_{\pm\textnormal{0.4}}$
\\

w/o HF
& $\textnormal{14.0}_{\pm\textnormal{2.6}}$ & $\textnormal{88.3}_{\pm\textnormal{1.5}}$
& $\textnormal{54.2}_{\pm\textnormal{0.9}}$ & $\textnormal{70.8}_{\pm\textnormal{0.5}}$
& $\textnormal{83.0}_{\pm\textnormal{0.2}}$ & $\textnormal{30.9}_{\pm\textnormal{0.5}}$
& $\textnormal{72.4}_{\pm\textnormal{0.2}}$ & $\textnormal{95.7}_{\pm\textnormal{0.1}}$ & $\textnormal{59.6}_{\pm\textnormal{0.3}}$
& $\textnormal{78.3}_{\pm\textnormal{0.2}}$ & $\textnormal{93.1}_{\pm\textnormal{0.1}}$ & $\textnormal{62.3}_{\pm\textnormal{0.3}}$
\\

\bottomrule
\end{tabular}
\end{table*}

\vspace{-0.2cm}
\begin{figure*}[h]
    \centering
    \begin{subfigure}[b]{0.195\textwidth}
        \centering
        \includegraphics[width=\textwidth]{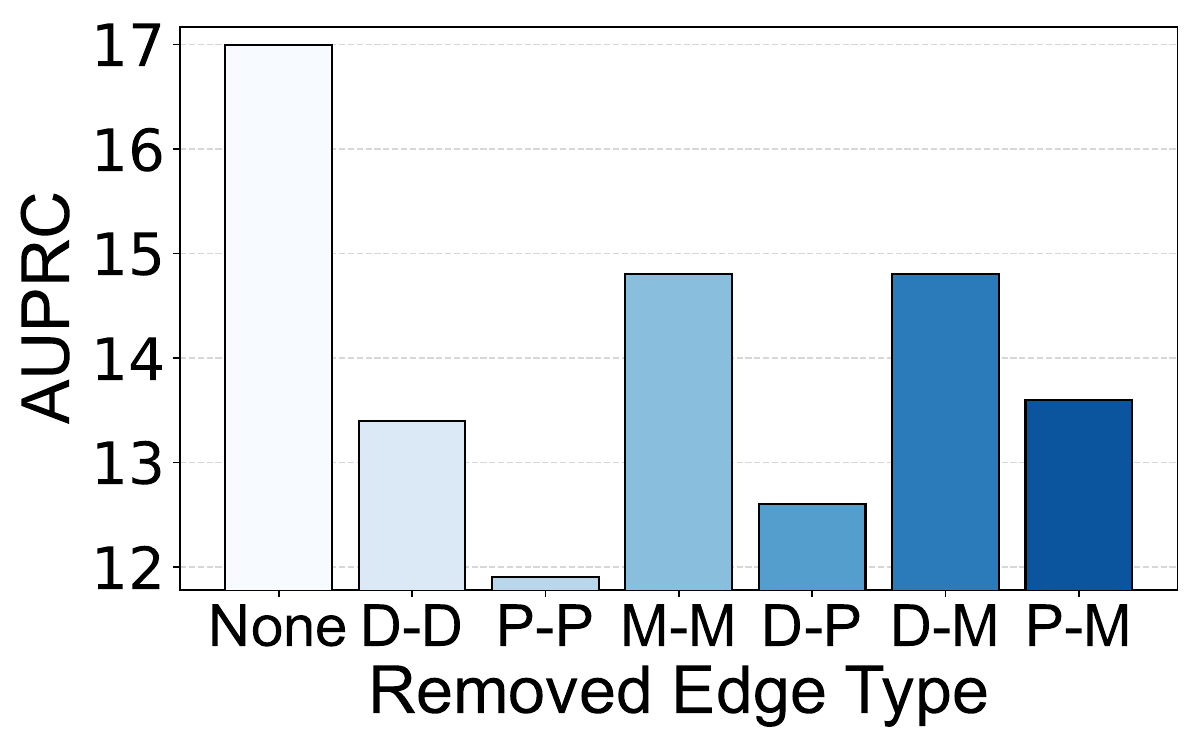}
        \caption{Mortality}
    \end{subfigure}
    \begin{subfigure}[b]{0.195\textwidth}
        \centering
        \includegraphics[width=\textwidth]{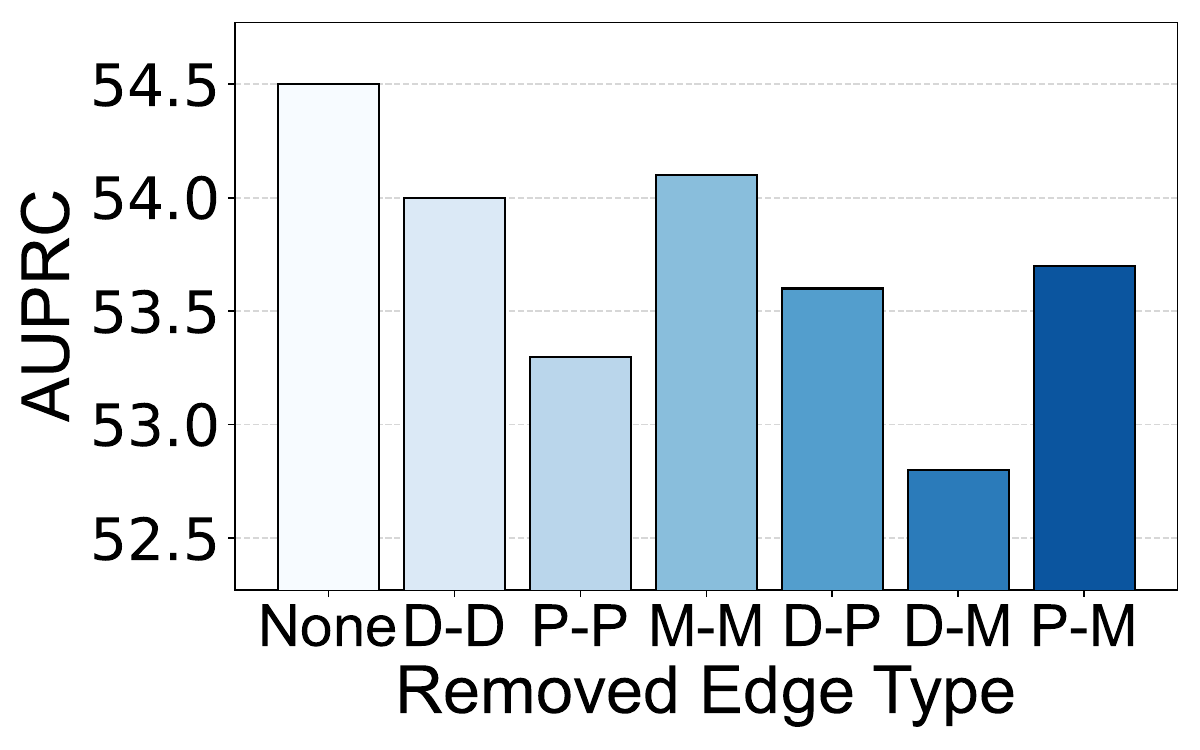}
        \caption{Readmission}
    \end{subfigure}
    \begin{subfigure}[b]{0.195\textwidth}
        \centering
        \includegraphics[width=\textwidth]{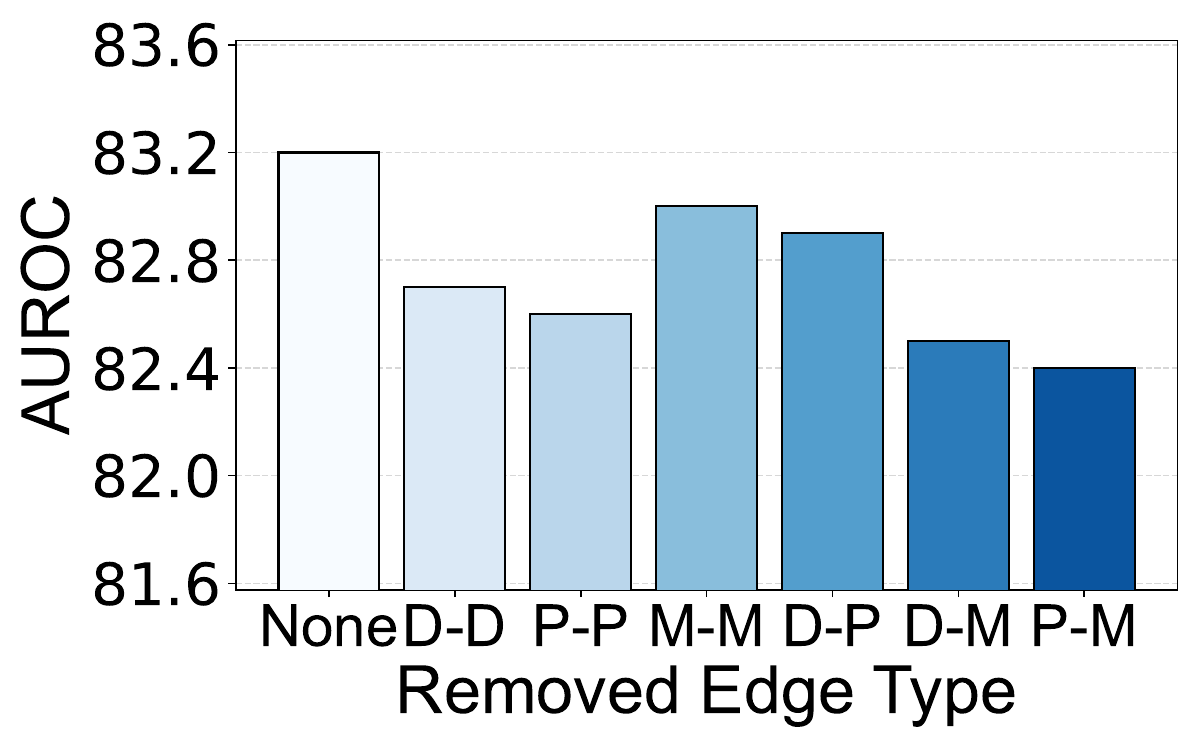}
        \caption{LoS}
    \end{subfigure}
    \begin{subfigure}[b]{0.195\textwidth}
        \centering
        \includegraphics[width=\textwidth]{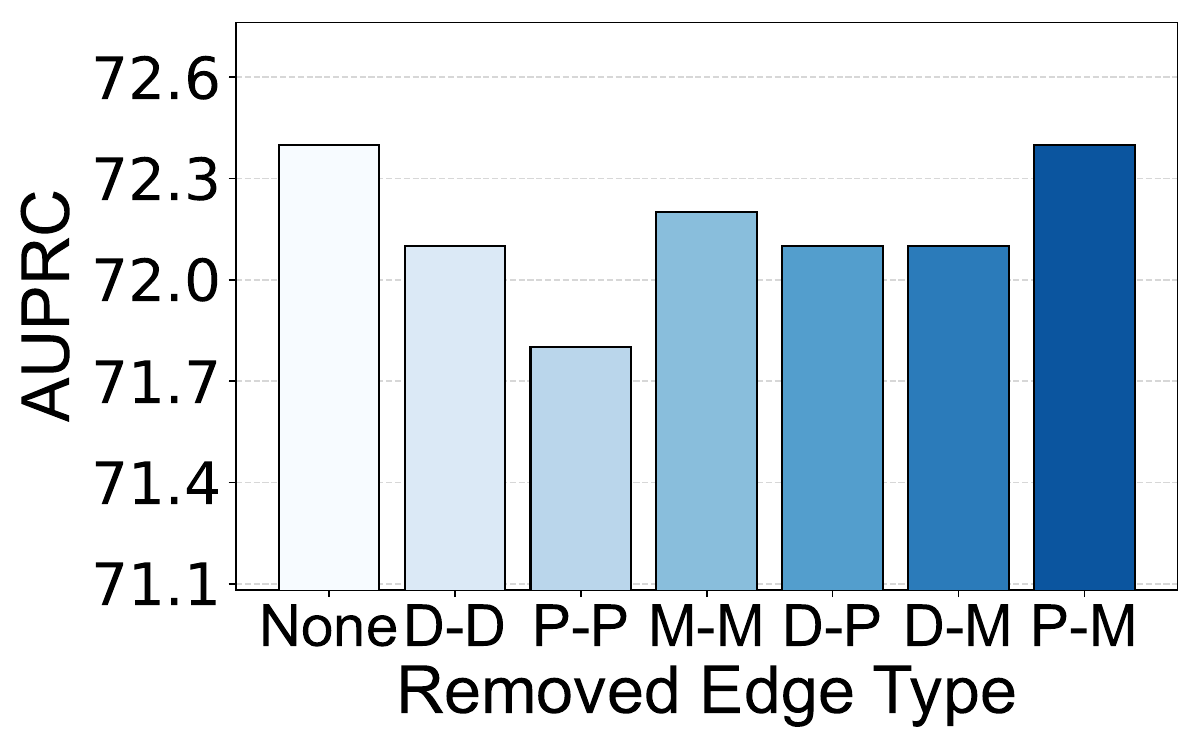}
        \caption{Drug}
    \end{subfigure}
    \begin{subfigure}[b]{0.195\textwidth}
        \centering
        \includegraphics[width=\textwidth]{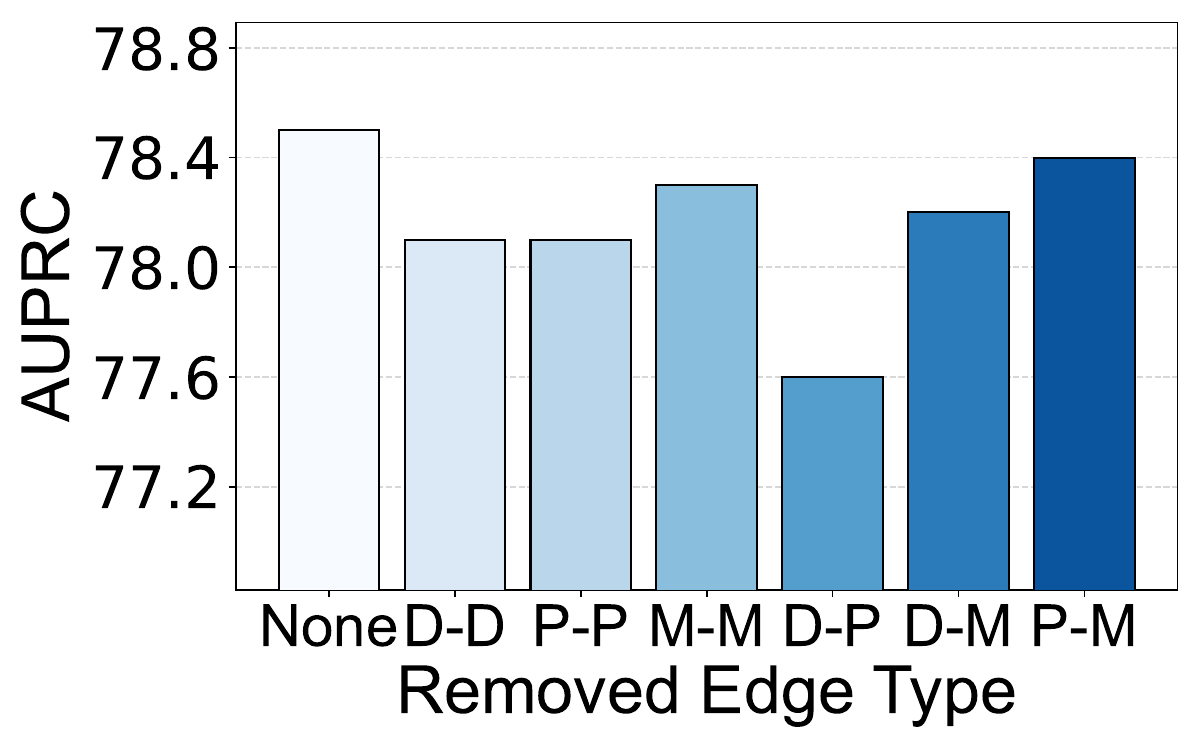}
        \caption{Phenotype}
    \end{subfigure}
    \vspace{-10px}
    \caption{Bar plots of model performance for five tasks on MIMIC-IV when ablation is performed on the six different types of edges in the medical KG. D, P, and M are the abbreviations for diagnosis, procedure, and medication/prescription respectively.}
    \label{fig:edge_rem}
    \vspace{-0.2cm}
\end{figure*}

\subsection{Ablation Study}
\label{sec:abl}

Detailed ablations are performed to analyze the contribution of the medical KG constructed, as well as the prototype learning modules. Five sets of ablations are performed, including ProtoEHR without medical KG, without code-, visit-, and patient-level learnable prototypes, and without the hierarchical fusion (HF) module. As shown in Table~\ref{tab:abl}, removing any of the components from the model results in poorer performance. From the results, we can see that removing the learned medical KG causes a 47.6\% decrease in AUPRC for the prediction of mortality. This is clear evidence that using medical KG to enhance the modeling within the code-level is critical for the prediction of the EHR and that our KG construction process is effective. In addition, removing code-, visit-, and patient-level prototypes results in a decrease of 5.3\%, 20.6\%, and 20.0\% AUPRC for the task, providing evidence that capturing intrinsic similarities within each level also benefits predictions.

As the importance of the medical KG is clear, we further perform ablations of the different types of edges in the KG to analyze individual contributions from the medical relations. Figures~\ref{fig:edge_rem}a to \ref{fig:edge_rem}e visualize the results of edge removal in MIMIC-IV. We can observe that removing $\text{P} \leftrightarrow \text{P}$ edges has the most noticeable effect compared to removing the edges of a single type of code. This is particularly significant for the prediction of mortality, resulting in a decrease in AUPRC 30\%. Procedures often indicate the severity of the condition, this knowledge is strengthened by linking between procedures, since procedures are generally applied simultaneously. Among the experiments that remove the edges between different types of codes, the removal of $\text{D} \leftrightarrow \text{P}$ connections has the greatest impact on three of the five tasks. Connections between diagnosis and procedure codes reveal the motivations behind the procedures. Tasks such as phenotype prediction require this knowledge to infer the diagnoses behind the visit, it is therefore reasonable for the model's predictive ability to decrease after detaching said edges. 

\subsection{Interpretability Study}
\label{sec:interp}
The interpretability granted by hierarchical prototype learning in our framework is two-fold: understanding which level of the hierarchy contributes the most to the predictions and the patterns that the hierarchical prototypes capture. We conducted these experiments in MIMIC-IV because it contains more patients, thereby producing more stable and robust results.

\subsubsection{\textbf{Modeling Level Importance v.s. Tasks}}
\label{sec:hf_interp}
\begin{figure}[tbp]
    \includegraphics[width=.9\linewidth]{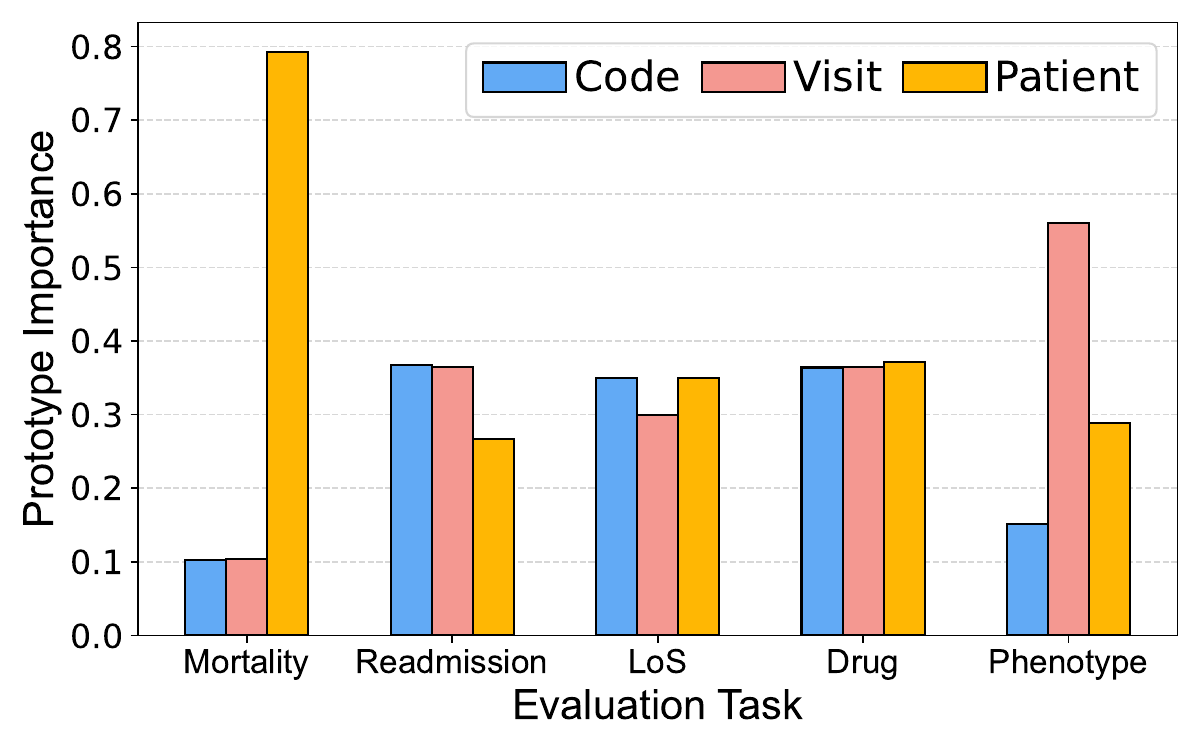}
    \vspace{-10px}
    \caption{Bar plots showing the contribution of code-, visit-, and patient-level prototype knowledge in the hierarchical fusion module for all five tasks on MIMIC-IV.}
    \label{fig:hf}
    \vspace{-15px}
\end{figure}

As explained in Section~\ref{sec:hf}, the fusion weights $\{\beta_t| t \in \{c,v,p\}\}$ represent the extent to which prototypes at each level contribute to the final representation and evaluation task. Hence, by recording the fusion weights for each sample in the test set, we can plot the average weights for each task, as shown in Figure~\ref{fig:hf}. To begin with, we can see that patient-level prototypes are the most important for mortality prediction. This is intuitive as mortality prediction is a patient-level task, it requires analyzing the health status of the patient as a whole to determine whether the patient will be deceased within the time frame, and the inclusion of patient-level cohort information improves this assessment. Moreover, despite drug recommendation and phenotype prediction being code-level tasks, the level each task deems important is different, most noticeably the contributions from the visit-level for phenotype prediction. We hypothesize that this is because diagnoses between visits are more correlated than drugs between visits. To verify this statement, we calculate the Jaccard score between the diagnoses/drugs of each patient's visit and the task label, and subsequently average the scores across patients in the test set. The values obtained are 0.287 and 0.439 for drug recommendation and phenotype prediction, respectively, supporting the fact that visit-level information is more important for phenotype prediction and therefore contributes more to predictions.

\subsubsection{\textbf{Prototype Importance Visualization}}
\begin{figure}[tbp]
    \includegraphics[width=1\linewidth]{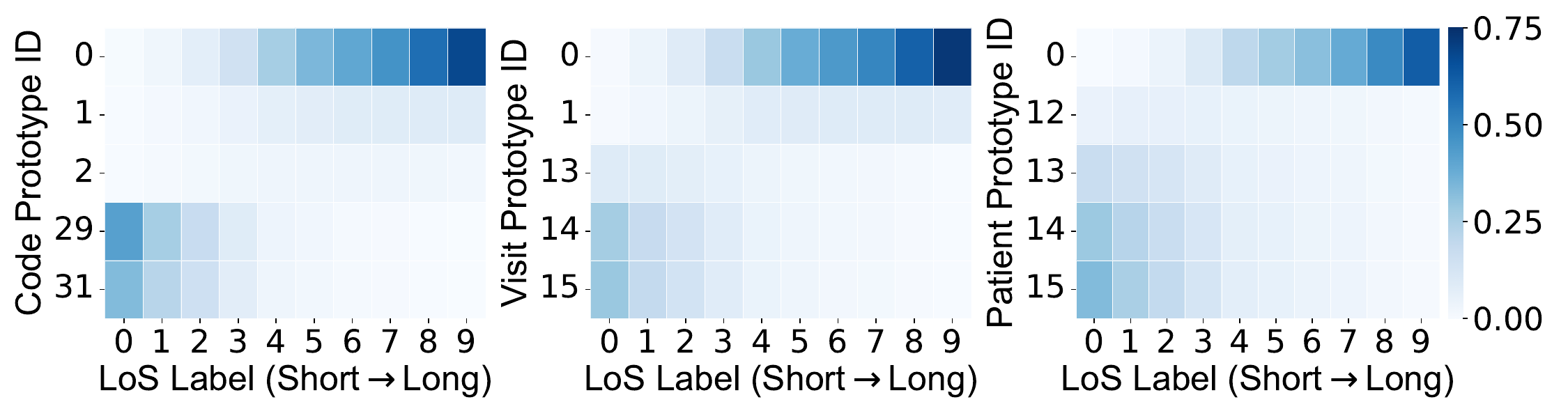}
    \vspace{-10px}
    \caption{Heat maps of code-, visit, and patient-level prototype importance against the length of stay task label on MIMIC-IV. The five most important prototypes are displayed.}
    \label{fig:hf_proto}
    \vspace{-10px}
\end{figure}

By analyzing the information prototypes encode, we  gain further insights into how prototypes encode intrinsic similarity within each level to enhance prediction. Figure~\ref{fig:hf_proto} visualizes the importance of each prototype in cross-attention for the prediction of the length of stay, where we selected the top five most important prototypes for analysis. For all three levels, as the task label increases (longer stay), some prototypes increase, whilst others decrease in significance in a monotonic manner. This suggests that different cohort knowledge is encoded by different prototypes, and for different patients, different cohort knowledge is used for prediction.

\begin{table}[htbp]
\caption{Top-5 diagnosis and top-5 procedure codes associated with the top-300 patients sorted by patient-level prototype importance for length of stay prediction on MIMIC-IV. First row: ID 0; Second row: ID 13.}
\label{tab:proto_top_codes}
\vspace{-10px}
\small
\begin{tabular}{|l|l|}
\hline
\textbf{Diagnosis Codes} & \textbf{Procedure Codes} \\ \hline

\begin{minipage}[t]{0.2\textwidth}
\begin{itemize}[itemsep=.05pt, leftmargin=5px]
  \item Residual Codes; Unclassified
  \item Essential Hypertension
  \item Disorders of Lipid Metabolism
  \item Complications of Surgical Procedures or Medical Care
  \item Other Aftercare
\end{itemize}
\end{minipage} &
\begin{minipage}[t]{0.24\textwidth}
\begin{itemize}[itemsep=.02pt, leftmargin=5px]
  \item Other Vascular Catheterization; \\Not Heart
  \item Extracorporeal Circulation Auxiliary \\to Open Heart Procedures
  \item Enteral and Parenteral Nutrition
  \item Coronary Artery Bypass Graft (CABG)
  \item Other OR Procedures on Vessels Other \\than Head and Neck
\end{itemize}
\end{minipage}
\\ \midrule

\begin{minipage}[t]{0.2\textwidth}
\begin{itemize}[itemsep=.02pt, leftmargin=5px]
  \item Essential Hypertension
  \item Nonspecific Chest Pain
  \item Residual Codes; Unclassified
  \item Alcohol-related Disorders
  \item Cardiac Dysrhythmias
\end{itemize}
\end{minipage} &
\begin{minipage}[t]{0.24\textwidth}
\begin{itemize}[itemsep=.02pt, leftmargin=5px]
  \item Other Diagnostic Procedures
  \item Routine Chest X-ray
  \item Electrocardiogram
  \item Other Therapeutic Procedures
  \item Other CT Scan
\end{itemize}
\end{minipage}
\\ \hline
\end{tabular}
\vspace{-0.3cm}
\end{table}

\subsubsection{\textbf{Prototype Effectiveness}}
Looking at the top 5 most frequent diagnosis and procedure codes present in the last visit of the top 300 patients of each patient-level prototype, the information that these prototypes capture could be inferred. The full results can be seen in Table~\ref{tab:proto_top_codes}. The importance of patient prototype ID 0 increases from 0.0034 to 0.62 as the patient's length of stay becomes longer, which can be validated by looking at the recorded medical codes. \texttt{Complications of Surgical Procedures or Medical Care}, \texttt{Other Vascular Catheterization; Not Heart}, and \texttt{Extracorporeal Circulation Auxiliary to Open Heart Procedures} are examples of these critical medical codes, generally requiring longer stays as the condition is severe and recovery is gradual. In contrast, since the importance of patient prototype ID 13 decreases from 0.18 to 0.0097 with increasing stay duration, the associated codes should represent less severe medical codes that typically require shorter hospital stays. As observed in the results, \texttt{Nonspecific Chest Pain}, \texttt{Alcohol-related Disorders}, and \texttt{Routine Chest X-ray} are related to this prototype. Conditions without complications are generally less complex, whilst results from routine checkups are often examined later without the need for the patient to stay, suggesting that this prototype captures medical events with earlier discharge. In a clinical setting, this interpretable characteristic of the model allows clinicians to pinpoint the medical codes/conditions that may have resulted in the deterioration of the patient's health. From another point of view, if a certain prototype repeatedly occurs for different patients, hospital management can inspect the associated codes to prevent the occurrence of any device malfunction or oversight.

To further quantify the effectiveness of the learned prototypes, we perform a clustering experiment to test whether the learned prototypes can be used to discover shared patterns in patients. For each patient in the test set, we retrieve three importance vectors of prototypes from the hierarchical fusion module—one corresponding to each hierarchy level—with each vector encoding the relative importance of that level’s prototypes (e.g., a visit-level vector of length four reflects the attention weights of the four learned visit-level prototypes). Subsequently, we apply K-Means clustering to each set of prototype vectors. The silhouette scores are computed for each cluster to measure the compactness and separation of the resulting clusters. Knowing that our learned prototypes are positively correlated with the task labels, higher silhouette scores would indicate that the prototypes effectively capture intrinsic similarities among patients and form a clinical pattern that is beneficial for prediction, thereby demonstrating their quantitative efficacy. The results obtained for mortality and length-of-stay prediction on MIMIC-IV are detailed in Table~\ref{tab:proto_sil}. In accord with the previous analysis in Figure~\ref{fig:hf_proto}, patient-level prototypes play an important role in mortality prediction, and three levels of prototypes contribute equally to length-of-stay prediction.

\begin{table}[h]
\vspace{-10px}
\caption{Silhouette scores between prototypes of each level for mortality and length-of-stay prediction tasks.}
\vspace{-10px}
\label{tab:proto_sil}
\begin{tabular}{l|ccc}
\toprule
\multicolumn{1}{c|}{\textbf{}} & \multicolumn{1}{c}{\textbf{Code}} & \multicolumn{1}{c}{\textbf{Visit}} & \multicolumn{1}{c}{\textbf{Patient}} \\ \midrule
\textbf{Mortality}      & 0.49 & 0.73 & 0.81 \\
\textbf{Length-of-Stay} & 0.56 & 0.56 & 0.54 \\
\bottomrule
\end{tabular}
\vspace{-10px}
\end{table}

\section{Related Work}
\paragraph{\emph{\textbf{EHR-based Healthcare Predictions.}}} Existing studies develop various models for EHR-based healthcare predictions. Specifically, GRAM \cite{choi2017gram} uses recurrent neural networks (RNNs) to capture the temporal dynamics of patient representations, while Deepr \cite{nguyen2016deepr} and AdaCare \cite{ma2020adacare} apply convolutional neural networks (CNNs) for visit information aggregation. StageNet \cite{gao2020stagenet} and MiME \cite{choi2018mime} further leverage both RNN and CNN to encode and combine visit embeddings for final patient representations.

Recent studies leverage graph structures to capture the relations between medical codes and visits for patient representation learning. For example, GT-BEHRT \citep{poulain2024graph} designs a graph transformer to better capture implicit information in long visit sequences, with more robust patient representations obtained. G-BERT \citep{shang2019pre} constructs a medical ontology tree to account for the relations between medical codes. Moreover, GraphCare \cite{jiang2023graphcare}, KerPrint \cite{yang2023kerprint} and SeqCare \cite{xu2023seqcare} either construct personalized KGs or introduce external KGs as the foundation for hierarchical representation learning, where learned representations are subsequently used for healthcare predictions. To further enhance the patient representations, GRASP \cite{zhang2021grasp} and PRISM \cite{zhu2024prism} introduce the prototype learning at the patient level only and learn patient-specific prototypes for sharing information.

Due to the high-stakes nature of healthcare, offering interpretability for EHR-based healthcare predictions endows trust in both patients and clinicians \cite{liu2025survunc}. MedPath \cite{ye2021medpath} offers interpretability by encapsulating disease progression paths on the KG that connect symptoms to diagnoses. KARE \cite{jiang2024reasoning} first augments the patient’s EHR context using a KG, and then leverages an LLM to analyze the augmented context for interpretable reasoning chains. GraphCare \cite{poulain2024graph} computes attention weights for nodes in a KG to capture their respective importance. Instead of using KGs, MedRetriever \cite{ye2021medretriever} extracts patient-relevant segments directly from the unstructured text, which serve as the rationale for model predictions.

The aforementioned studies underscore the importance of explicitly modeling the hierarchical structure of EHR data, while also highlighting the need to capture intrinsic similarities therein. In contrast to approaches that focus solely on patient-level similarity, our method systematically incorporates intrinsic similarities at all levels of the hierarchy. We construct patient representations by progressively aggregating information from the code level, through the visit level, and ultimately to the patient level. Additionally, the learned prototypes capture clinically meaningful patterns, enhancing both predictive performance and interpretability.

\paragraph{\emph{\textbf{Prototype Learning.}}} Prototype learning refers to a set of machine learning techniques that identifies or extracts a representative ``prototype''  reflecting the overall information of data within a specific group. This approach has been widely adopted in computer vision. For instance, ClusterFormer \cite{liang2024clusterfomer} iteratively learns prototypes of image features using cross-attention to resemble expectation maximization. ProtoPNet \citep{chen2019looks} uses prototypes as weights for a CNN to highlight different properties of the images.  
ProtoGAN \citep{kumar2019protogan} acquires class prototypes by training a network to map class features to a lower-dimensional space. Another method \citep{gupta2023class} applies the moving average to update class prototypes to use as anchors for contrastive learning. Motivated by its advantage of capturing shared information, we introduce hierarchical prototype learning to capture the intrinsic similarity at all three levels, as well as directly enhancing the patient representation at the final fusion stage.

\section{Conclusion}
In this work, we proposed a novel and interpretable EHR prediction framework that takes advantage of prototype learning and hierarchical learning to explore the within-level similarities and cross-level hierarchy for healthcare prediction. Comprehensive experiments on mortality prediction, readmission prediction, length of stay prediction, drug recommendation, and phenotype prediction were conducted across two datasets to determine the effectiveness of the model. Our interpretability study supports the efficacy of using prototype learning to unravel the reasons behind model predictions, which includes examining the contributions of each level and the importance of hierarchical prototypes toward the output. 

In future work, we aim to enhance the construction and learning of medical KGs to better capture the correlations between medical codes and task labels, and to incorporate additional clinical information such as laboratory test results to further improve predictive performance while validating on a more diverse set of datasets.

\section{Acknowledgements}
Tingting Zhu was supported by the Royal Academy of Engineering under the Research Fellowship scheme.

\appendix

\section{Details of Medical KG Construction}
During the \textbf{retrieval} stage, we use an open-source LLM (Llama 3 70B \cite{dubey2024llama}) to extract candidate relations, yielding 398{,}555 triplets. As these include spurious links, a \textbf{cleaning} stage follows: we ask GPT-4 to label a subset of 30{,}000 triplets with binary validity, of which 14{,}378 are judged \emph{true}. We then train a classifier on this labeled subset by first embedding each triplet with BERT \cite{kenton2019bert} and feeding the embeddings to a multilayer perceptron (MLP) to predict validity. Applying the trained classifier to the remaining unlabeled triplets marks 122{,}422 as likely true at a probability threshold of $\geq 0.5$. For the final KG, we combine all 14{,}378 GPT-verified true triplets with the top 71{,}890 classifier-scored triplets (the highest-probability subset, sized at $5\times$ the GPT-verified positives), resulting in a KG with 86{,}268 triplets and 713 entities (including one for padding).

% During the \textbf{retrieval} stage, an open-source LLM (Llama3-70b \cite{dubey2024llama}) is used to extract triplets, which, however, contain several false connections and need to be removed to build an expressive and robust medical KG. Thus, in the \textbf{cleaning} stage, we used GPT-4 to verify a subset of the triplets by outputting binary assessments. A total of 30,000 triplets (14,378 of which are true) are verified out of the 398,555 triplets retrieved in the first stage, using for training the classifier. We first embed each triplet using BERT \cite{kenton2019bert}, and the word embeddings are passed through the multilayer perceptron (MLP) network to output binary classifications. This process further reduces the number of triplets, leaving 122,422 remaining, with a probability that each triplet is true $\geq 0.5$. Our final KG contains 713 nodes (including one for padding) and 86,268 triplets: 14,378 triplets are verified by GPT and 71,890 additional, most confident triplets are selected by the classifier trained on the GPT-verified data.

The aim of the final \textbf{refinement} stage is to reduce the number of unique edges to obtain generalized representations of medical knowledge. We apply agglomerative clustering with ward linkage to merge edges based on their word embedding distance. This method is effective in recognizing edges with similar names. However, this also introduces the problem that semantically opposing edges could differ by a single word, but could still be grouped as similar. Hence, we use the same LLM as in the second stage to determine whether any cluster contains semantically opposite edges. Triplet refinement reduces the number of unique edges from 2,330 to merely 269, significantly improving the robustness of the constructed medical KG. Note that the directed edges are subsequently reversed for the CompGCN layer. All the prompt designs are provided in the implementation codes.

\section{LLM Reliability for KG Construction}
\label{sec:appendix_c}
We assess the reliability of our medical KG constructed using LLMs, focusing on its efficiency, stability, and robustness.

Our approach exhaustively retrieves candidate relations by querying each code pair using an LLM. This offline procedure ensures comprehensive coverage without incurring repeated computational costs. The resulting KG centers on general semantic relations (e.g., ``causes''), forming a strong foundation for downstream healthcare tasks. To address concerns about the reliability of LLM-generated facts, we implemented a multi-stage filtering and refinement process to eliminate low-confidence or implausible triplets. These steps contribute to a high-quality, consistent, and non-redundant KG.

In our framework, Llama-70B is used to extract relations between medical code pairs. We evaluated its stability by randomly selecting 100 codes from the set of medical codes $\mathcal{C}$ to generate all possible code pairs and extract the linking relation five times for each pair. We then embedded these relations using BERT and computed cosine similarities between the five extractions for each pair. A mean and a standard deviation were obtained for cosine similarities within the five extractions. The overall mean cosine similarity across all code pairs was 0.93 with a standard deviation of only 0.05. These results demonstrate that Llama produces highly consistent outputs, and any sampling randomness is negligible. Compared to traditional KG construction, which relies heavily on manual curation by domain experts, our LLM-based strategy offers a scalable and cost-effective alternative for clinically meaningful graphs.

\section{LLM for Direct Healthcare Predictions}

In addition to the compared baselines, one might question the effectiveness of using LLMs directly for healthcare prediction tasks. Here, we use prompts directly with Llama models to perform mortality prediction on MIMIC-III, which are shown in Table~\ref{tab:llm_mort-table}. Both Llama-8B and Llama-70B are unsuitable for directly performing mortality prediction when compared to the F1-score achieved by ProtoEHR of 49.1. The smaller 8B model produces invalid outputs at a much higher rate compared to the 70B model, highlighting that prompt engineering alone is insufficient to mitigate these issues in smaller models. Even though the 70B model performs slightly better, its prediction reliability remains unsatisfactory, and the associated deployment costs further discourage its practical application. From the Llama output, we also observe that both models exhibit a bias that frequently predicts patient death, resulting in a high false positive rate. These results reinforce the necessity of our tailored, data-driven approach.

\begin{table}[htbp]
\caption{Performance of Llama for mortality prediction on MIMIC-III. T denotes temperature.}
\label{tab:llm_mort-table}
\vspace{-10px}
\centering
\begin{tabular}{c|c|c}
\toprule
\textbf{Model (T)} & \textbf{F1 Score} & \textbf{Error Rate} \\
\midrule
Llama-8B (T=0)   & 0.06      & 14.56\% \\
Llama-8B (T=0.7) & 0.05$\pm$0.01 & 25.61\% \\
Llama-70B (T=0)  & 0.08      & 0.09\%  \\
Llama-70B (T=0.7)& 0.07$\pm$0.00 & 3.91\%  \\
\bottomrule
\end{tabular}
\end{table}

\clearpage

\section*{GenAI Usage Disclosure}
LLMs are used during this research for two main purposes: (1) medical KG construction by proposing triplets between medical codes and determining whether the triplets are true or false, and (2) performing grammar and clarity checks during the preparation of the paper. The LLM-generated KG has been validated from multiple aspects, as shown in the experiment section and appendix. No text was generated without human review, and all research contributions are the result of the authors’ original work.

\bibliographystyle{ACM-Reference-Format}
\balance
\bibliography{references}


\begin{thebibliography}{38}

%%% the Web address.

\ifx \showCODEN    \undefined \def \showCODEN     #1{\unskip}     \fi
\ifx \showDOI      \undefined \def \showDOI       #1{#1}\fi
\ifx \showISBNx    \undefined \def \showISBNx     #1{\unskip}     \fi
\ifx \showISBNxiii \undefined \def \showISBNxiii  #1{\unskip}     \fi
\ifx \showISSN     \undefined \def \showISSN      #1{\unskip}     \fi
\ifx \showLCCN     \undefined \def \showLCCN      #1{\unskip}     \fi
\ifx \shownote     \undefined \def \shownote      #1{#1}          \fi
\ifx \showarticletitle \undefined \def \showarticletitle #1{#1}   \fi
\ifx \showURL      \undefined \def \showURL       {\relax}        \fi
% invisible to TeX
\providecommand\bibfield[2]{#2}
\providecommand\bibinfo[2]{#2}
\providecommand\natexlab[1]{#1}
\providecommand\showeprint[2][]{arXiv:#2}

\bibitem[Achiam et~al\mbox{.}(2023)]%
        {achiam2023gpt}
\bibfield{author}{\bibinfo{person}{Josh Achiam}, \bibinfo{person}{Steven Adler}, \bibinfo{person}{Sandhini Agarwal}, \bibinfo{person}{Lama Ahmad}, \bibinfo{person}{Ilge Akkaya}, \bibinfo{person}{Florencia~Leoni Aleman}, \bibinfo{person}{Diogo Almeida}, \bibinfo{person}{Janko Altenschmidt}, \bibinfo{person}{Sam Altman}, \bibinfo{person}{Shyamal Anadkat}, {et~al\mbox{.}}} \bibinfo{year}{2023}\natexlab{}.
\newblock \showarticletitle{{GPT-4} technical report}.
\newblock \bibinfo{journal}{\emph{arXiv preprint arXiv:2303.08774}} (\bibinfo{year}{2023}).
\newblock


\bibitem[Akiba et~al\mbox{.}(2019)]%
        {akiba2019optuna}
\bibfield{author}{\bibinfo{person}{Takuya Akiba}, \bibinfo{person}{Shotaro Sano}, \bibinfo{person}{Toshihiko Yanase}, \bibinfo{person}{Takeru Ohta}, {and} \bibinfo{person}{Masanori Koyama}.} \bibinfo{year}{2019}\natexlab{}.
\newblock \showarticletitle{{O}ptuna: A Next-Generation Hyperparameter Optimization Framework}. In \bibinfo{booktitle}{\emph{The 25th ACM SIGKDD International Conference on Knowledge Discovery \& Data Mining}}. \bibinfo{pages}{2623--2631}.
\newblock


\bibitem[Bodenreider(2004)]%
        {bodenreider2004unified}
\bibfield{author}{\bibinfo{person}{Olivier Bodenreider}.} \bibinfo{year}{2004}\natexlab{}.
\newblock \showarticletitle{The unified medical language system (UMLS): integrating biomedical terminology}.
\newblock \bibinfo{journal}{\emph{Nucleic acids research}} \bibinfo{volume}{32}, \bibinfo{number}{suppl\_1} (\bibinfo{year}{2004}), \bibinfo{pages}{D267--D270}.
\newblock


\bibitem[Chen et~al\mbox{.}(2019)]%
        {chen2019looks}
\bibfield{author}{\bibinfo{person}{Chaofan Chen}, \bibinfo{person}{Oscar Li}, \bibinfo{person}{Daniel Tao}, \bibinfo{person}{Alina Barnett}, \bibinfo{person}{Cynthia Rudin}, {and} \bibinfo{person}{Jonathan~K Su}.} \bibinfo{year}{2019}\natexlab{}.
\newblock \showarticletitle{This looks like that: Deep learning for interpretable image recognition}.
\newblock \bibinfo{journal}{\emph{NeurIPS}}  \bibinfo{volume}{32} (\bibinfo{year}{2019}).
\newblock


\bibitem[Choi et~al\mbox{.}(2017)]%
        {choi2017gram}
\bibfield{author}{\bibinfo{person}{Edward Choi}, \bibinfo{person}{Mohammad~Taha Bahadori}, \bibinfo{person}{Le Song}, \bibinfo{person}{Walter~F Stewart}, {and} \bibinfo{person}{Jimeng Sun}.} \bibinfo{year}{2017}\natexlab{}.
\newblock \showarticletitle{GRAM: graph-based attention model for healthcare representation learning}. In \bibinfo{booktitle}{\emph{KDD}}. \bibinfo{pages}{787--795}.
\newblock


\bibitem[Choi et~al\mbox{.}(2016)]%
        {choi2016retain}
\bibfield{author}{\bibinfo{person}{Edward Choi}, \bibinfo{person}{Mohammad~Taha Bahadori}, \bibinfo{person}{Jimeng Sun}, \bibinfo{person}{Joshua Kulas}, \bibinfo{person}{Andy Schuetz}, {and} \bibinfo{person}{Walter Stewart}.} \bibinfo{year}{2016}\natexlab{}.
\newblock \showarticletitle{{RETAIN}: An interpretable predictive model for healthcare using reverse time attention mechanism}.
\newblock \bibinfo{journal}{\emph{NIPS}}  \bibinfo{volume}{29} (\bibinfo{year}{2016}).
\newblock


\bibitem[Choi et~al\mbox{.}(2018)]%
        {choi2018mime}
\bibfield{author}{\bibinfo{person}{Edward Choi}, \bibinfo{person}{Cao Xiao}, \bibinfo{person}{Walter Stewart}, {and} \bibinfo{person}{Jimeng Sun}.} \bibinfo{year}{2018}\natexlab{}.
\newblock \showarticletitle{{MiME}: Multilevel medical embedding of electronic health records for predictive healthcare}.
\newblock \bibinfo{journal}{\emph{NeurIPS}}  \bibinfo{volume}{31} (\bibinfo{year}{2018}).
\newblock


\bibitem[Dubey et~al\mbox{.}(2024)]%
        {dubey2024llama}
\bibfield{author}{\bibinfo{person}{Abhimanyu Dubey}, \bibinfo{person}{Abhinav Jauhri}, \bibinfo{person}{Abhinav Pandey}, \bibinfo{person}{Abhishek Kadian}, \bibinfo{person}{Ahmad Al-Dahle}, \bibinfo{person}{Aiesha Letman}, \bibinfo{person}{Akhil Mathur}, \bibinfo{person}{Alan Schelten}, \bibinfo{person}{Amy Yang}, \bibinfo{person}{Angela Fan}, {et~al\mbox{.}}} \bibinfo{year}{2024}\natexlab{}.
\newblock \showarticletitle{The {Llama} 3 herd of models}.
\newblock \bibinfo{journal}{\emph{arXiv preprint arXiv:2407.21783}} (\bibinfo{year}{2024}).
\newblock


\bibitem[Gao et~al\mbox{.}(2020)]%
        {gao2020stagenet}
\bibfield{author}{\bibinfo{person}{Junyi Gao}, \bibinfo{person}{Cao Xiao}, \bibinfo{person}{Yasha Wang}, \bibinfo{person}{Wen Tang}, \bibinfo{person}{Lucas~M Glass}, {and} \bibinfo{person}{Jimeng Sun}.} \bibinfo{year}{2020}\natexlab{}.
\newblock \showarticletitle{{StageNet}: Stage-aware neural networks for health risk prediction}. In \bibinfo{booktitle}{\emph{WWW}}. \bibinfo{pages}{530--540}.
\newblock


\bibitem[Gupta et~al\mbox{.}(2023)]%
        {gupta2023class}
\bibfield{author}{\bibinfo{person}{Rohit Gupta}, \bibinfo{person}{Anirban Roy}, \bibinfo{person}{Claire Christensen}, \bibinfo{person}{Sujeong Kim}, \bibinfo{person}{Sarah Gerard}, \bibinfo{person}{Madeline Cincebeaux}, \bibinfo{person}{Ajay Divakaran}, \bibinfo{person}{Todd Grindal}, {and} \bibinfo{person}{Mubarak Shah}.} \bibinfo{year}{2023}\natexlab{}.
\newblock \showarticletitle{Class prototypes based contrastive learning for classifying multi-label and fine-grained educational videos}. In \bibinfo{booktitle}{\emph{CVPR}}. \bibinfo{pages}{19923--19933}.
\newblock


\bibitem[Harutyunyan et~al\mbox{.}(2019)]%
        {harutyunyan2019multitask}
\bibfield{author}{\bibinfo{person}{Hrayr Harutyunyan}, \bibinfo{person}{Hrant Khachatrian}, \bibinfo{person}{David~C Kale}, \bibinfo{person}{Greg Ver~Steeg}, {and} \bibinfo{person}{Aram Galstyan}.} \bibinfo{year}{2019}\natexlab{}.
\newblock \showarticletitle{Multitask learning and benchmarking with clinical time series data}.
\newblock \bibinfo{journal}{\emph{Scientific Data}} \bibinfo{volume}{6}, \bibinfo{number}{1} (\bibinfo{year}{2019}), \bibinfo{pages}{96}.
\newblock


\bibitem[Jiang et~al\mbox{.}(2024a)]%
        {jiang2023graphcare}
\bibfield{author}{\bibinfo{person}{Pengcheng Jiang}, \bibinfo{person}{Cao Xiao}, \bibinfo{person}{Adam Cross}, {and} \bibinfo{person}{Jimeng Sun}.} \bibinfo{year}{2024}\natexlab{a}.
\newblock \showarticletitle{{GraphCare}: Enhancing healthcare predictions with personalized knowledge graphs}.
\newblock \bibinfo{journal}{\emph{ICLR}} (\bibinfo{year}{2024}).
\newblock


\bibitem[Jiang et~al\mbox{.}(2024b)]%
        {jiang2024reasoning}
\bibfield{author}{\bibinfo{person}{Pengcheng Jiang}, \bibinfo{person}{Cao Xiao}, \bibinfo{person}{Minhao Jiang}, \bibinfo{person}{Parminder Bhatia}, \bibinfo{person}{Taha Kass-Hout}, \bibinfo{person}{Jimeng Sun}, {and} \bibinfo{person}{Jiawei Han}.} \bibinfo{year}{2024}\natexlab{b}.
\newblock \showarticletitle{Reasoning-enhanced healthcare predictions with knowledge graph community retrieval}.
\newblock \bibinfo{journal}{\emph{arXiv preprint arXiv:2410.04585}} (\bibinfo{year}{2024}).
\newblock


\bibitem[Johnson et~al\mbox{.}(2023)]%
        {johnson2020mimic}
\bibfield{author}{\bibinfo{person}{Alistair~EW Johnson}, \bibinfo{person}{Lucas Bulgarelli}, \bibinfo{person}{Lu Shen}, \bibinfo{person}{Alvin Gayles}, \bibinfo{person}{Ayad Shammout}, \bibinfo{person}{Steven Horng}, \bibinfo{person}{Tom~J Pollard}, \bibinfo{person}{Sicheng Hao}, \bibinfo{person}{Benjamin Moody}, \bibinfo{person}{Brian Gow}, {et~al\mbox{.}}} \bibinfo{year}{2023}\natexlab{}.
\newblock \showarticletitle{{MIMIC-IV}, a freely accessible electronic health record dataset}.
\newblock \bibinfo{journal}{\emph{Scientific Data}} \bibinfo{volume}{10}, \bibinfo{number}{1} (\bibinfo{year}{2023}), \bibinfo{pages}{1}.
\newblock


\bibitem[Johnson et~al\mbox{.}(2016)]%
        {johnson2016mimic}
\bibfield{author}{\bibinfo{person}{Alistair~EW Johnson}, \bibinfo{person}{Tom~J Pollard}, \bibinfo{person}{Lu Shen}, \bibinfo{person}{Li-wei~H Lehman}, \bibinfo{person}{Mengling Feng}, \bibinfo{person}{Mohammad Ghassemi}, \bibinfo{person}{Benjamin Moody}, \bibinfo{person}{Peter Szolovits}, \bibinfo{person}{Leo Anthony~Celi}, {and} \bibinfo{person}{Roger~G Mark}.} \bibinfo{year}{2016}\natexlab{}.
\newblock \showarticletitle{{MIMIC-III}, a freely accessible critical care database}.
\newblock \bibinfo{journal}{\emph{Scientific Data}} \bibinfo{volume}{3}, \bibinfo{number}{1} (\bibinfo{year}{2016}), \bibinfo{pages}{1--9}.
\newblock


\bibitem[Kannel et~al\mbox{.}(1961)]%
        {kannel1961factors}
\bibfield{author}{\bibinfo{person}{William~B Kannel}, \bibinfo{person}{Thomas~R Dawber}, \bibinfo{person}{Abraham Kagan}, \bibinfo{person}{Nicholas Revotskie}, {and} \bibinfo{person}{Joseph Stokes~III}.} \bibinfo{year}{1961}\natexlab{}.
\newblock \showarticletitle{Factors of risk in the development of coronary heart disease—six-year follow-up experience: the Framingham Study}.
\newblock \bibinfo{journal}{\emph{Annals of Internal Medicine}} \bibinfo{volume}{55}, \bibinfo{number}{1} (\bibinfo{year}{1961}), \bibinfo{pages}{33--50}.
\newblock


\bibitem[Kenton and Toutanova(2019)]%
        {kenton2019bert}
\bibfield{author}{\bibinfo{person}{Jacob Devlin Ming-Wei~Chang Kenton} {and} \bibinfo{person}{Lee~Kristina Toutanova}.} \bibinfo{year}{2019}\natexlab{}.
\newblock \showarticletitle{Bert: Pre-training of deep bidirectional transformers for language understanding}. In \bibinfo{booktitle}{\emph{NAACL}}, Vol.~\bibinfo{volume}{1}. Minneapolis, Minnesota.
\newblock


\bibitem[Kumar~Dwivedi et~al\mbox{.}(2019)]%
        {kumar2019protogan}
\bibfield{author}{\bibinfo{person}{Sai Kumar~Dwivedi}, \bibinfo{person}{Vikram Gupta}, \bibinfo{person}{Rahul Mitra}, \bibinfo{person}{Shuaib Ahmed}, {and} \bibinfo{person}{Arjun Jain}.} \bibinfo{year}{2019}\natexlab{}.
\newblock \showarticletitle{{ProtoGAN}: Towards few shot learning for action recognition}. In \bibinfo{booktitle}{\emph{ICCV Workshop}}. \bibinfo{pages}{0--0}.
\newblock


\bibitem[Liang et~al\mbox{.}(2024)]%
        {liang2024clusterfomer}
\bibfield{author}{\bibinfo{person}{James Liang}, \bibinfo{person}{Yiming Cui}, \bibinfo{person}{Qifan Wang}, \bibinfo{person}{Tong Geng}, \bibinfo{person}{Wenguan Wang}, {and} \bibinfo{person}{Dongfang Liu}.} \bibinfo{year}{2024}\natexlab{}.
\newblock \showarticletitle{{ClusterFomer}: clustering as a universal visual learner}.
\newblock \bibinfo{journal}{\emph{NeurIPS}}  \bibinfo{volume}{36} (\bibinfo{year}{2024}).
\newblock


\bibitem[Liu et~al\mbox{.}(2025)]%
        {liu2025survunc}
\bibfield{author}{\bibinfo{person}{Yu Liu}, \bibinfo{person}{Weiyao Tao}, \bibinfo{person}{Tong Xia}, \bibinfo{person}{Simon Knight}, {and} \bibinfo{person}{Tingting Zhu}.} \bibinfo{year}{2025}\natexlab{}.
\newblock \showarticletitle{SurvUnc: A meta-model based uncertainty quantification framework for survival analysis}. In \bibinfo{booktitle}{\emph{KDD}}. \bibinfo{pages}{1903--1914}.
\newblock


\bibitem[Ma et~al\mbox{.}(2020)]%
        {ma2020adacare}
\bibfield{author}{\bibinfo{person}{Liantao Ma}, \bibinfo{person}{Junyi Gao}, \bibinfo{person}{Yasha Wang}, \bibinfo{person}{Chaohe Zhang}, \bibinfo{person}{Jiangtao Wang}, \bibinfo{person}{Wenjie Ruan}, \bibinfo{person}{Wen Tang}, \bibinfo{person}{Xin Gao}, {and} \bibinfo{person}{Xinyu Ma}.} \bibinfo{year}{2020}\natexlab{}.
\newblock \showarticletitle{{AdaCare}: Explainable clinical health status representation learning via scale-adaptive feature extraction and recalibration}. In \bibinfo{booktitle}{\emph{AAAI}}, Vol.~\bibinfo{volume}{34}. \bibinfo{pages}{825--832}.
\newblock


\bibitem[Nguyen et~al\mbox{.}(2016)]%
        {nguyen2016deepr}
\bibfield{author}{\bibinfo{person}{Phuoc Nguyen}, \bibinfo{person}{Truyen Tran}, \bibinfo{person}{Nilmini Wickramasinghe}, {and} \bibinfo{person}{Svetha Venkatesh}.} \bibinfo{year}{2016}\natexlab{}.
\newblock \showarticletitle{Deepr: A convolutional net for medical records}.
\newblock \bibinfo{journal}{\emph{JBHI}} \bibinfo{volume}{21}, \bibinfo{number}{1} (\bibinfo{year}{2016}), \bibinfo{pages}{22--30}.
\newblock


\bibitem[Poulain and Beheshti(2024)]%
        {poulain2024graph}
\bibfield{author}{\bibinfo{person}{Raphael Poulain} {and} \bibinfo{person}{Rahmatollah Beheshti}.} \bibinfo{year}{2024}\natexlab{}.
\newblock \showarticletitle{Graph transformers on EHRs: Better representation improves downstream performance}. In \bibinfo{booktitle}{\emph{ICLR}}.
\newblock


\bibitem[Rokach and Maimon(2005)]%
        {rokach2005clustering}
\bibfield{author}{\bibinfo{person}{Lior Rokach} {and} \bibinfo{person}{Oded Maimon}.} \bibinfo{year}{2005}\natexlab{}.
\newblock \showarticletitle{Clustering methods}.
\newblock \bibinfo{journal}{\emph{Data Mining and Knowledge Discovery Handbook}} (\bibinfo{year}{2005}), \bibinfo{pages}{321--352}.
\newblock


\bibitem[Sarwar et~al\mbox{.}(2022)]%
        {sarwar2022secondary}
\bibfield{author}{\bibinfo{person}{Tabinda Sarwar}, \bibinfo{person}{Sattar Seifollahi}, \bibinfo{person}{Jeffrey Chan}, \bibinfo{person}{Xiuzhen Zhang}, \bibinfo{person}{Vural Aksakalli}, \bibinfo{person}{Irene Hudson}, \bibinfo{person}{Karin Verspoor}, {and} \bibinfo{person}{Lawrence Cavedon}.} \bibinfo{year}{2022}\natexlab{}.
\newblock \showarticletitle{The secondary use of electronic health records for data mining: Data characteristics and challenges}.
\newblock \bibinfo{journal}{\emph{CSUR}} \bibinfo{volume}{55}, \bibinfo{number}{2} (\bibinfo{year}{2022}), \bibinfo{pages}{1--40}.
\newblock


\bibitem[Shang et~al\mbox{.}(2019a)]%
        {shang2019pre}
\bibfield{author}{\bibinfo{person}{Junyuan Shang}, \bibinfo{person}{Tengfei Ma}, \bibinfo{person}{Cao Xiao}, {and} \bibinfo{person}{Jimeng Sun}.} \bibinfo{year}{2019}\natexlab{a}.
\newblock \showarticletitle{Pre-training of graph augmented transformers for medication recommendation}. In \bibinfo{booktitle}{\emph{IJCAI}}.
\newblock


\bibitem[Shang et~al\mbox{.}(2019b)]%
        {shang2019gamenet}
\bibfield{author}{\bibinfo{person}{Junyuan Shang}, \bibinfo{person}{Cao Xiao}, \bibinfo{person}{Tengfei Ma}, \bibinfo{person}{Hongyan Li}, {and} \bibinfo{person}{Jimeng Sun}.} \bibinfo{year}{2019}\natexlab{b}.
\newblock \showarticletitle{Gamenet: Graph augmented memory networks for recommending medication combination}. In \bibinfo{booktitle}{\emph{AAAI}}, Vol.~\bibinfo{volume}{33}. \bibinfo{pages}{1126--1133}.
\newblock


\bibitem[Shickel et~al\mbox{.}(2017)]%
        {shickel2017deep}
\bibfield{author}{\bibinfo{person}{Benjamin Shickel}, \bibinfo{person}{Patrick~James Tighe}, \bibinfo{person}{Azra Bihorac}, {and} \bibinfo{person}{Parisa Rashidi}.} \bibinfo{year}{2017}\natexlab{}.
\newblock \showarticletitle{Deep EHR: a survey of recent advances in deep learning techniques for electronic health record (EHR) analysis}.
\newblock \bibinfo{journal}{\emph{JBHI}} \bibinfo{volume}{22}, \bibinfo{number}{5} (\bibinfo{year}{2017}), \bibinfo{pages}{1589--1604}.
\newblock


\bibitem[Vashishth et~al\mbox{.}(2020)]%
        {vashishth2020composition}
\bibfield{author}{\bibinfo{person}{Shikhar Vashishth}, \bibinfo{person}{Soumya Sanyal}, \bibinfo{person}{Vikram Nitin}, {and} \bibinfo{person}{Partha Talukdar}.} \bibinfo{year}{2020}\natexlab{}.
\newblock \showarticletitle{Composition-based multi-relational graph convolutional networks}. In \bibinfo{booktitle}{\emph{ICLR}}.
\newblock


\bibitem[Wang et~al\mbox{.}(2020)]%
        {wang2020mimic}
\bibfield{author}{\bibinfo{person}{Shirly Wang}, \bibinfo{person}{Matthew~BA McDermott}, \bibinfo{person}{Geeticka Chauhan}, \bibinfo{person}{Marzyeh Ghassemi}, \bibinfo{person}{Michael~C Hughes}, {and} \bibinfo{person}{Tristan Naumann}.} \bibinfo{year}{2020}\natexlab{}.
\newblock \showarticletitle{{MIMIC-Extract}: A data extraction, preprocessing, and representation pipeline for {MIMIC-III}}. In \bibinfo{booktitle}{\emph{CHIL}}. \bibinfo{pages}{222--235}.
\newblock


\bibitem[Wang and Zhu(2021)]%
        {wang2021predictive}
\bibfield{author}{\bibinfo{person}{Shuwen Wang} {and} \bibinfo{person}{Xingquan Zhu}.} \bibinfo{year}{2021}\natexlab{}.
\newblock \showarticletitle{Predictive modeling of hospital readmission: challenges and solutions}.
\newblock \bibinfo{journal}{\emph{TCCB}} \bibinfo{volume}{19}, \bibinfo{number}{5} (\bibinfo{year}{2021}), \bibinfo{pages}{2975--2995}.
\newblock


\bibitem[Xu et~al\mbox{.}(2023)]%
        {xu2023seqcare}
\bibfield{author}{\bibinfo{person}{Yongxin Xu}, \bibinfo{person}{Xu Chu}, \bibinfo{person}{Kai Yang}, \bibinfo{person}{Zhiyuan Wang}, \bibinfo{person}{Peinie Zou}, \bibinfo{person}{Hongxin Ding}, \bibinfo{person}{Junfeng Zhao}, \bibinfo{person}{Yasha Wang}, {and} \bibinfo{person}{Bing Xie}.} \bibinfo{year}{2023}\natexlab{}.
\newblock \showarticletitle{{SeqCare:} Sequential training with external medical knowledge graph for diagnosis prediction in healthcare data}. In \bibinfo{booktitle}{\emph{WWW}}. \bibinfo{pages}{2819--2830}.
\newblock


\bibitem[Yang et~al\mbox{.}(2023)]%
        {yang2023kerprint}
\bibfield{author}{\bibinfo{person}{Kai Yang}, \bibinfo{person}{Yongxin Xu}, \bibinfo{person}{Peinie Zou}, \bibinfo{person}{Hongxin Ding}, \bibinfo{person}{Junfeng Zhao}, \bibinfo{person}{Yasha Wang}, {and} \bibinfo{person}{Bing Xie}.} \bibinfo{year}{2023}\natexlab{}.
\newblock \showarticletitle{KerPrint: Local-global knowledge graph enhanced diagnosis prediction for retrospective and prospective interpretations}. In \bibinfo{booktitle}{\emph{AAAI}}, Vol.~\bibinfo{volume}{37}. \bibinfo{pages}{5357--5365}.
\newblock


\bibitem[Ye et~al\mbox{.}(2021a)]%
        {ye2021medpath}
\bibfield{author}{\bibinfo{person}{Muchao Ye}, \bibinfo{person}{Suhan Cui}, \bibinfo{person}{Yaqing Wang}, \bibinfo{person}{Junyu Luo}, \bibinfo{person}{Cao Xiao}, {and} \bibinfo{person}{Fenglong Ma}.} \bibinfo{year}{2021}\natexlab{a}.
\newblock \showarticletitle{Medpath: Augmenting health risk prediction via medical knowledge paths}. In \bibinfo{booktitle}{\emph{WWW}}. \bibinfo{pages}{1397--1409}.
\newblock


\bibitem[Ye et~al\mbox{.}(2021b)]%
        {ye2021medretriever}
\bibfield{author}{\bibinfo{person}{Muchao Ye}, \bibinfo{person}{Suhan Cui}, \bibinfo{person}{Yaqing Wang}, \bibinfo{person}{Junyu Luo}, \bibinfo{person}{Cao Xiao}, {and} \bibinfo{person}{Fenglong Ma}.} \bibinfo{year}{2021}\natexlab{b}.
\newblock \showarticletitle{Medretriever: Target-driven interpretable health risk prediction via retrieving unstructured medical text}. In \bibinfo{booktitle}{\emph{CIKM}}. \bibinfo{pages}{2414--2423}.
\newblock


\bibitem[Zhang et~al\mbox{.}(2021)]%
        {zhang2021grasp}
\bibfield{author}{\bibinfo{person}{Chaohe Zhang}, \bibinfo{person}{Xin Gao}, \bibinfo{person}{Liantao Ma}, \bibinfo{person}{Yasha Wang}, \bibinfo{person}{Jiangtao Wang}, {and} \bibinfo{person}{Wen Tang}.} \bibinfo{year}{2021}\natexlab{}.
\newblock \showarticletitle{GRASP: Generic framework for health status representation learning based on incorporating knowledge from similar patients}. In \bibinfo{booktitle}{\emph{AAAI}}, Vol.~\bibinfo{volume}{35}. \bibinfo{pages}{715--723}.
\newblock


\bibitem[Zhu et~al\mbox{.}(2024a)]%
        {zhu2024llms}
\bibfield{author}{\bibinfo{person}{Yuqi Zhu}, \bibinfo{person}{Xiaohan Wang}, \bibinfo{person}{Jing Chen}, \bibinfo{person}{Shuofei Qiao}, \bibinfo{person}{Yixin Ou}, \bibinfo{person}{Yunzhi Yao}, \bibinfo{person}{Shumin Deng}, \bibinfo{person}{Huajun Chen}, {and} \bibinfo{person}{Ningyu Zhang}.} \bibinfo{year}{2024}\natexlab{a}.
\newblock \showarticletitle{LLMs for knowledge graph construction and reasoning: Recent capabilities and future opportunities}.
\newblock \bibinfo{journal}{\emph{WWW}} \bibinfo{volume}{27}, \bibinfo{number}{5} (\bibinfo{year}{2024}), \bibinfo{pages}{58}.
\newblock


\bibitem[Zhu et~al\mbox{.}(2024b)]%
        {zhu2024prism}
\bibfield{author}{\bibinfo{person}{Yinghao Zhu}, \bibinfo{person}{Zixiang Wang}, \bibinfo{person}{Long He}, \bibinfo{person}{Shiyun Xie}, \bibinfo{person}{Xiaochen Zheng}, \bibinfo{person}{Liantao Ma}, {and} \bibinfo{person}{Chengwei Pan}.} \bibinfo{year}{2024}\natexlab{b}.
\newblock \showarticletitle{PRISM: Mitigating EHR Data sparsity via learning from missing feature calibrated prototype patient representations}. In \bibinfo{booktitle}{\emph{CIKM}}. \bibinfo{pages}{3560--3569}.
\newblock


\end{thebibliography}

\end{document}